\documentclass[sigconf]{aamas} 

\usepackage{makecell}
\usepackage{booktabs}

\usepackage{balance} 
\usepackage{multirow}
\usepackage{tcolorbox}
\usepackage{enumitem}
\tcbuselibrary{listings,breakable}
\usepackage{subcaption}

\setcopyright{ifaamas}
\acmConference[AAMAS '26]{Proc.\@ of the 25th International Conference
on Autonomous Agents and Multiagent Systems (AAMAS 2026)}{May 25 -- 29, 2026}
{Paphos, Cyprus}{C.~Amato, L.~Dennis, V.~Mascardi, J.~Thangarajah (eds.)}
\copyrightyear{2026}
\acmYear{2026}
\acmDOI{}
\acmPrice{}
\acmISBN{}

\acmSubmissionID{487}

\title[AAMAS-2026 Formatting Instructions]{R-Debater: Retrieval-Augmented Debate Generation through Argumentative Memory}

\author{Maoyuan Li}
\authornote{Equal contribution.}
\affiliation{
  \institution{Wuhan College of Communication}
  \city{}
  \country{}
}
\email{mli302@aucklanduni.ac.nz}

\author{Zhongsheng Wang}
\authornotemark[1]
\affiliation{
  \institution{Wuhan College of Communication}
  \city{University of Auckland}
  \country{}
}
\email{zhongsheng.wang@auckland.ac.nz}

\author{Haoyuan Li}
\affiliation{
  \institution{University of Auckland}
  \city{}
  \country{}
}
\email{hli962@aucklanduni.ac.nz}

\author{Jiamou Liu}
\affiliation{
  \institution{University of Auckland}
  \city{}
  \country{}
}
\email{jiamou.liu@auckland.ac.nz}

\begin{abstract}
We present R-Debater, an agentic framework for generating multi-turn debates built on argumentative memory. Grounded in rhetoric and memory studies, the system views debate as a process of recalling and adapting prior arguments to maintain stance consistency, respond to opponents, and support claims with evidence. Specifically, R-Debater integrates a debate knowledge base for retrieving case-like evidence and prior debate moves with a role-based agent that composes coherent utterances across turns. We evaluate on standardized ORCHID debates, constructing a 1,000-item retrieval corpus and a held-out set of 32 debates across seven domains. Two tasks are evaluated: next-utterance generation, assessed by InspireScore (subjective, logical, and factual), and adversarial multi-turn simulations, judged by Debatrix (argument, source, language, and overall). Compared with strong LLM baselines, R-Debater achieves higher single-turn and multi-turn scores. Human evaluation with 20 experienced debaters further confirms its consistency and evidence use, showing that combining retrieval grounding with structured planning yields more faithful, stance-aligned, and coherent debates across turns.
Code and supplementary materials are available at \url{https://anonymous.4open.science/r/R-debater-E87F/}.

\end{abstract}


\newcommand{\BibTeX}{\rm B\kern-.05em{\sc i\kern-.025em b}\kern-.08em\TeX}

\begin{document}


\pagestyle{fancy}
\fancyhead{}


\maketitle 


\section{Introduction}
Competitive debate distills public reasoning into a structured, adversarial setting where speakers must plan multi-turn rhetoric, maintain stance, and ground claims in verifiable evidence. Prior work, from end-to-end debating systems (e.g., \emph{Project Debater} \cite{slonim2021autonomous}) to a mature literature on argument mining \cite{lawrence-reed-2019-argument}, has advanced pipelines for discovering and organizing arguments. 
Meanwhile, Large Language Models (LLMs) have markedly improved open‐ended dialogue and text generation~\cite{qin2024large,kumar2024large}, but when faced with competitive debate, their outputs often remain fluent yet shallow: They are insufficiently grounded, weak on stance fidelity, and brittle over many turns. Therefore, one should pivot to the capacities that make argumentation stance-faithful and durable across turns.

Building on this insight, we treat argumentation not only as a computational task but as a cognitive and rhetorical process grounded in human memory and discourse. As Vitale notes, public argumentation operates through a \emph{rhetorical-argumentative memory} that recycles and reformulates prior persuasive strategies in new situations~\cite{vitale2015rhetorical}. Likewise, Aleida Assmann in her acclaimed theory of cultural memory conceptualizes collective memory as a dialogic, dynamic reconstruction of past discourses rather than a static archive~\cite{assmann2015dialogic}. These perspectives motivate our central premise: \emph{effective debate generation should retrieve and recontextualize argumentative memory, producing statements that are grounded in prior reasoning and robust rhetorical patterns}.

\begin{figure}[t]
    \centering
    \includegraphics[width=\columnwidth]{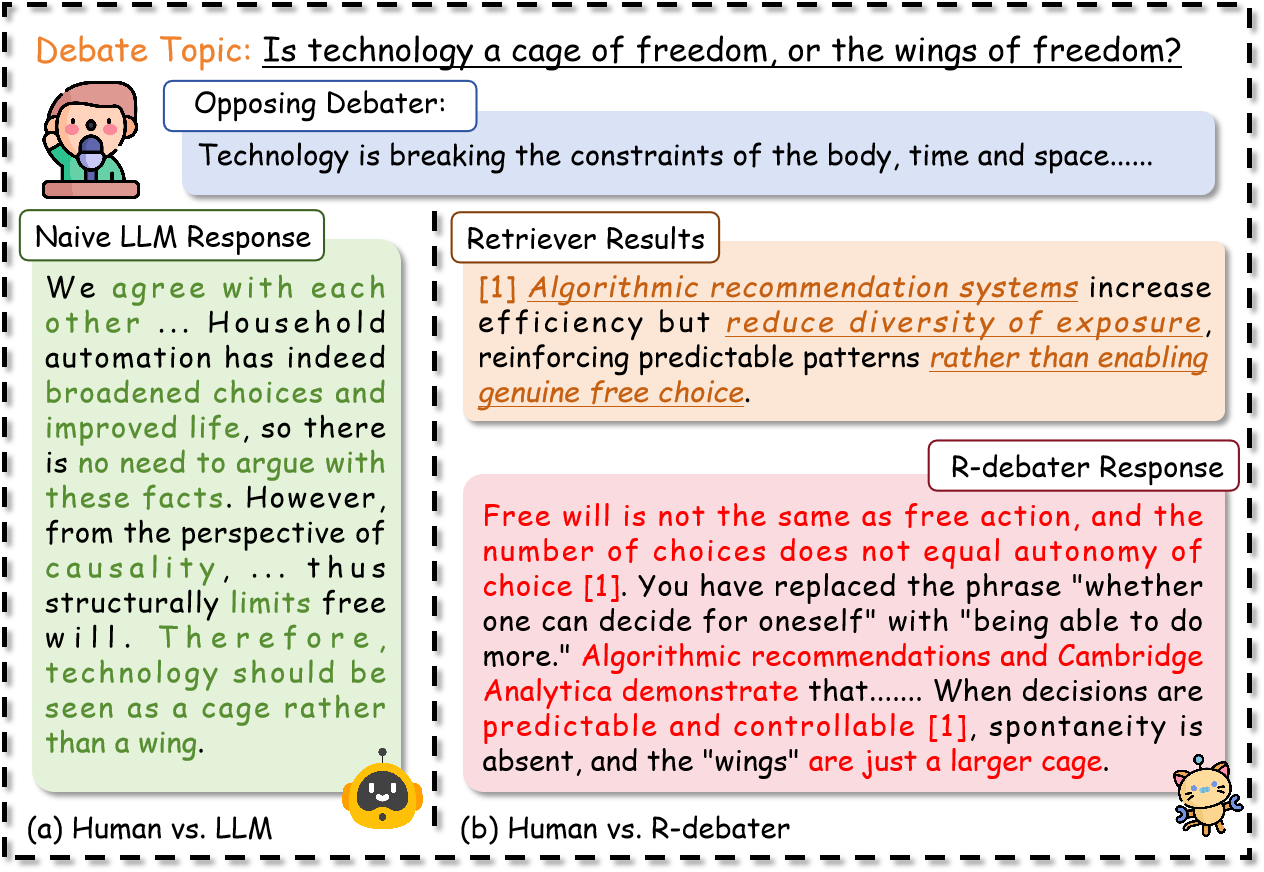}
    \caption{Humans debate by integrating external knowledge and debate strategies, while LLMs often fail to connect retrieval with argumentation. Our task aims to bridge this gap by enhancing the ability of LLMs to utilize materials grounded in retrieved debate history and strategies more effectively.}
    \label{fig:debate_example}
\end{figure}

Retrieval‐augmented generation (RAG) is a natural mechanism for recalling prior cases and evidence~\cite{sharma2024og,lewis2020retrieval}, and has proved useful in QA and open‐domain assistants~\cite{shuster2021blenderbot2}. 
However, mainstream RAG stacks optimize for short factual responses and struggle with debate's structured, adversarial recontextualization, where models must reconcile retrieved material with evolving discourse and opponent claims. 
Recent analyses also show a persistent ``tug-of-war'' between a model's internal priors and retrieved evidence, yielding faithfulness and coverage errors without explicit control~\cite{howfaithfulrag2024,zhang2025raghallucination}. 
Concurrently, agentic AI frameworks provide machinery for planning and tool use~\cite{wu2024autogen,li2023camel}, yet they rarely encode argumentative schemes~\cite{walton2008argumentation} or dialogic memory needed to preserve stance and reasoning coherence across turns \cite{zhang2024can}.


This paper addresses this gap by conceptualizing
argumentative memory in computational terms.
Specifically, we treat retrieval as the mechanism for recalling prior discourse and role-based planning as the mechanism for reframing it into new argumentative settings.
Tackling this question raises three central challenges:
First, generated statements must incorporate appropriate argumentative schemes and logical structures to be persuasive and coherent.
Second, the system must maintain stance fidelity throughout the debate, ensuring that rebuttals are grounded in opponents’ claims while avoiding hallucinations or sycophantic tendencies.
Finally, the model must retrieve high-quality, context-relevant debate materials and integrate them strategically into statement generation rather than reproducing retrieved content verbatim.

To address these challenges, we propose \textbf{R-Debater}, a framework that integrates retrieval-augmented reasoning with role-based planning for debate statement generation. Unlike prior systems, R-Debater leverages the entire debate history to generate stance-specific and rhetorically coherent utterances, bridging the gap between rhetorical theory and generative modeling.

We conduct extensive experiments on the ORCHID~\cite{zhao2024orchid} dataset, which comprises over 1,000 formal debates across multiple domains. 
Evaluation demonstrates that R-Debater achieves substantial gains in both logical fidelity and rhetorical persuasiveness over strong LLM and RAG baselines. 
In particular, it delivers near-perfect logical coherence and markedly higher factual grounding, while human experts prefer its generated debates in over \textbf{75\%} of pairwise comparisons. 
These results confirm that R-Debater not only improves measurable debate quality but also aligns closely with expert judgements, validating its reliability and interpretability.

Overall, our main contributions are as follows:
\begin{itemize}[leftmargin=*]
    \item We present, to the best of our knowledge, the first systematic investigation of debate statement generation with LLMs under realistic multi-turn settings.
    \item We propose R-Debater, a novel framework that integrates argumentation structures, debate strategies, stance fidelity mechanisms, and retrieval-based case reasoning to address the core challenges of debate generation.
    \item Extensive empirical validation on debate datasets demonstrates that R-Debater significantly outperforms strong LLM and RAG baselines in terms of factual accuracy, stance consistency, and cross-turn coherence.
\end{itemize}


\section{Related Work}
\subsection{Computational Argumentation}

Argumentation research has long-standing roots in symbolic reasoning and formal logic~\cite{van2004systematic,van2003development,van1988rationale,eemeren2018argumentation,van2016argumentation,wachsmuth2018argumentation,el2019computational}, with early computational approaches offering transparency and rigor but relying on brittle, rule-based knowledge bases that limit scalability to open-domain debates~\cite{zukerman2000using,d2005value,charwat2015methods,toledo2016expert}. 
These symbolic methods also often lack expressiveness in realistic discourse.

To address these limitations, subsequent work in computational argumentation has leveraged advances in natural language processing (NLP)~\cite{Castagna2024ComputationalAC,lawrence2020argument,guerraoui2023teach,vassiliades2021argumentation}. Retrieval-based pipelines such as CANDELA~\cite{hua2019argument,hua2018neural} adopt a retrieval-planning-realization process, aggregating evidence from multiple sources to improve the factual grounding and specificity. 
Other studies enhance neural models with argumentation knowledge graphs~\cite{al2021employing}, which encode structured relationships between claims, premises, and stances, thereby improving both credibility and topical relevance. 
Beyond single-turn generation, hierarchical dialogue frameworks~\cite{sato2015end,sakai2020hierarchical} integrate topic analysis, stance retrieval, and user feedback to produce more adaptive multi-turn interactions. 
These advances substantially enhance fluency, evidence utilization, and contextual relevance.
However, most existing approaches still treat each debate as an isolated reasoning task rather than a reactivation of prior argumentative patterns. They focus on structural modeling but lack cross-turn continuity, making it difficult to capture genuine debate dynamics.

Recent studies highlight that LLMs possess strong capabilities in long-context understanding and coherent discourse generation~\cite{an2024make,jin2024llm,lin2024infinite,ding2024longrope}. These advances open the door to debate systems that move beyond turn-local reasoning and model the holistic argumentative flow. Building on this line, R-Debater aims to integrate external knowledge with full-dialogue modeling to produce strategically grounded and rhetorically coherent argumentative responses.

\subsection{Retrieval-Augmented and Agentic AI}

LLMs have demonstrated strong long-text generation capabilities across diverse NLP tasks~\cite{qin2024large,kumar2024large,zhang2024comprehensive,Zhao2023ASO}, yet they remain prone to hallucination and lack the factual grounding required for credible debate-oriented applications~\cite{Huang2023ASO}. 
Retrieval-augmented generation (RAG) mitigates these issues by grounding outputs in external knowledge sources~\cite{Lewis2020RetrievalAugmentedGF,Gao2023RetrievalAugmentedGF}, and RAG-based dialogue systems~\cite{Li2024RetrievalAG,Mao2020GenerationAugmentedRF,Li2021ConversationsAN} have shown improved factuality and coherence. 
However, these systems primarily target informational QA or open-domain chit-chat~\cite{Zhao2024LongRAGAD,Chen2023BenchmarkingLL}, failing to address the complex requirements of stance construction and rebuttal across multiple debate turns. 
In this sense, RAG can be regarded as an early computational analogue of \emph{rhetorical memory}, capable of retrieving and reorganizing prior discourses to support new argumentative goals, yet still lacking the dynamic recontextualization central to real debates.

Parallel to RAG, research on agentic AI explores decomposing complex tasks into interacting role-based agents. 
Frameworks such as AutoGen~\cite{Wu2023AutoGenEN} and Agent4Debate~\cite{feng2024m} demonstrate that specialized agents can coordinate, critique, or compete to achieve more robust and strategically informed outputs~\cite{Zhao2024LongRAGAD,Chen2023BenchmarkingLL}, but the absence of explicit evidence grounding often limits their argumentative credibility. 
Viewed through the lens of dialogic memory, such systems simulate interactional structure yet fail to capture the cumulative recall of argumentation history.

In contrast, R-Debater builds on both paradigms by explicitly operationalizing \emph{argumentative memory} through retrieval-augmented and role-based mechanisms. 
It dynamically retrieves debate-specific evidence, assigns strategic roles, and recontextualizes past argumentative moves to generate text that is both stance-specific and globally coherent in rhetorical and strategic senses, thereby achieving a deeper integration of evidence retrieval and rhetorical reasoning within complex multi-turn debates.

\section{Task Formulation}

\label{sec:Formulation}

\text{Dialogue-based argumentation.} A debate is a structured, adversarial dialogue in which two parties alternate turns to advance their own stance while challenging the opponent. 
Unlike open-domain dialogues, a debate imposes strict constraints: utterances must be \emph{stance-consistent} and \emph{argumentatively relevant}, either reinforcing one's position or rebutting opposing claims.

Formally, we represent a {\em (dialogue-based) argumentation} $A$ as a pair $A=(T,H)$, where 
$T=\{t_1,t_2\}$ denotes the two opposing stances (pro and con) on a given topic, and 
the ordered sequence $$H=(u_1,u_2,\dots,u_n)$$ represents an $n$-turn argument history (where $n \in \mathbb{N}$ is even). 
Here, $u_j$ denotes the utterance at turn $j$, which is a natural-language text. 
We assume the pro side speaks first, so the set $P_n=(u_j)_{j\equiv1\pmod{2}}=(u_1,u_3,\ldots,u_{n-1})$ contains all utterances produced by the pro side, 
while $C_n=(u_j)_{j\equiv0\pmod{2}}=(u_2,u_4,\ldots,u_n)$ contains utterances produced by the con side.


We introduce a single-utterance predicate $\phi(u,s)$ that returns \texttt{True} if utterance $u$ either explicitly supports stance $s$ or rebuts the opposing stance. We require that in the argument history $H$, every utterance $u_i$ either satisfies $\phi(u_i,t_1)$ if $i$ is odd, or satisfies $\phi(u_i,t_2)$ if $i$ is even, but not both. 



\smallskip

\textbf{Debate and argumentative memory.}  From the perspective of rhetorical and dialogic memory \cite{vitale2015rhetorical}, each utterance not only presents a local claim but also reconstructs argumentative contexts to sustain stance coherence across turns. We thus view a debate as an iterative reconstruction of prior reasoning contexts rather than a sequence of isolated utterances: At each turn, a speaker retrieves and adapts fragments of earlier arguments, evidence, or rhetorical strategies to sustain coherence and respond to the opponent's reasoning.
We refer to this process as \emph{argumentative memory}.
We now formulate a process in which argumentative memory is instrumental the construction of utterane in a debate.

We assume to have a pre-curated set $\mathcal{D}$ of latent memory of previously-encountered argumentative patterns. 
Given an ongoing (partial) dialogue history $H_j=(u_1,\ldots,u_j)$, where $j<n$ and stances $T$, the debater retrieves a subset of relevant memory items $\mathcal{E}_j$ and synthesizes a stance-consistent next utterance $u_{j+1}$:
\begin{equation}\label{eqn:problem}
\mathcal{E}_j = \mathcal{R}(H_j, \mathcal{D}), \,
u_{j+1} = \mathcal{M}(T, H_j, \mathcal{E}_j).
\end{equation}
where $\mathcal{R}$ denotes the {\em retrieval} function that recalls an early memory $\mathcal{E}_j$, 
and $\mathcal{M}$ denotes the {\em reasoning} function that adapts $\mathcal{E}_j$ to the current debate context to produce the next utterance $u_{j+1}$. 

\smallskip

\textbf{R-Debater Task Definition.} We define debate utterance generation as a \emph{next-utterance prediction} problem that requires stance fidelity and argumentative relevance while leveraging argumentative memory and rhetorical strategies.
Suppose that we have an argumentative memory set $\mathcal{D}$. 
Given stances $T$, partial history $H_j$ with even $j<n$, and opponent $\mathcal{C}$, 
the problem seeks a pair of retrieval and reasoning functions $(R,M)$ that allows us (i.e., Pros $\mathcal{P}$) to generate a valid argumentation with the opponent. In other words, applying $R$ and $M$ repeatedly to produce the sequence of utterances $H=(u_1,u_2,\ldots,u_n)$, where each utterance in $C_n=(u_2,u_4,\ldots)$ is produced by the opponent $\mathcal{C}$, and $u_{i+1}$ is produced by $\mathcal{M}(T, H_i, \mathcal{R}(H_i, \mathcal{D}))$ for even $i\geq j$. Then the pair $(T,H)$ meets the definition of an argumentation above, i.e., $\phi(u_{i+1}, s) = \texttt{True}$ for all even $i\geq j$.

\section{Methodology}

\begin{figure*}[!t]
  \centering
  \includegraphics[width=\textwidth]{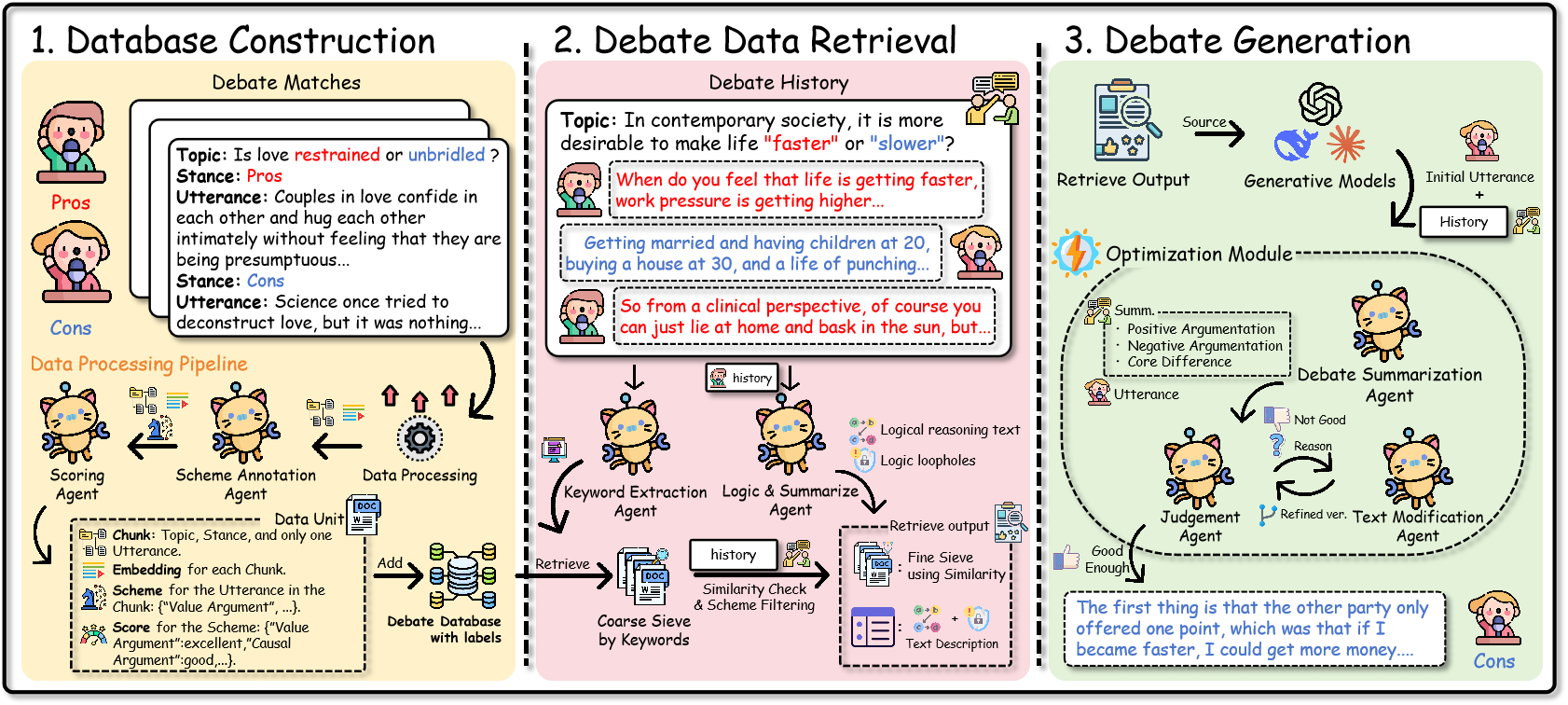}
  \caption{
    R-Debater consists of three modules: 
    (1) Database Construction, 
    (2) Debate Data Retrieval, and 
    (3) Debate Generation. 
    Together, they enable the dynamic reconstruction of argumentative memory for generating next-turn debates.
    }
  \label{fig:big-figure}
\end{figure*}

\label{sec:method}
We propose \textbf{R-Debater}, as illustrated in Figure~\ref{fig:big-figure}, a retrieval-
augmented and role-based framework that models debate as the dynamic reconstruction of argumentative memory. 
It consists of three pipelines: 
(1) \textbf{Database Construction}, which segments debate transcripts into utterance-level units and annotates argumentative information, 
(2) \textbf{Debate Data Retrieval}, which leverages this database to obtain stance-relevant evidence while also summarizing the logical gaps and weaknesses in the opponent's reasoning, and 
(3) \textbf{Debate Generation}, which produces stance-consistent, rhetorically coherent utterances through iterative verification.

\subsection{Database Construction}
\label{sec:DatabaseConstruction}

We construct a large-scale debate knowledge base from authentic transcripts collected across diverse public sources. 
Each transcript is processed by a rule-based parser that segments dialogue into utterance-level units $U=\{u_1,\dots,u_n\}$, 
removing moderator turns and meta-commentary to retain only argumentative content. 
Each $u_i$ corresponds to a single debater’s statement at a particular debate stage. 
A pre-trained encoder $f_{\text{emb}}$ then maps every utterance into a dense representation $\mathbf{e}_i=f_{\text{emb}}(u_i)$, 
yielding pairs $(u_i,\mathbf{e}_i)$ that capture semantic meaning and stylistic features for subsequent retrieval and reasoning.

\smallskip

\textbf{Argumentation Schemes.}
Following Walton et al.~\cite{walton2008argumentation}, an \textit{argumentation scheme} is a stereotypical reasoning pattern that shows how premises support a conclusion and poses critical questions for evaluating validity.
These schemes are widely used in computational argumentation to represent and assess the plausibility of natural arguments. 
As Walton et al. stress in their seminal monograph \textit{Argumentation Schemes}~\cite{walton2008argumentation}, schemes serve as dialogical templates that capture recurring structures of everyday reasoning and provide systematic criteria for evaluating strength.
In this work, we employ seven commonly adopted schemes from~\cite{walton2008argumentation}: 
\textit{example-based argumentation},
\textit{expert opinion},
\textit{positive consequence},
\textit{negative consequence},
\textit{causal argumentation},
\textit{analogical argumentation},
and \textit{value-based argumentation},
which together provide the reasoning foundation for our framework.

\textbf{Scheme Annotation and Scoring.}
The \textbf{Scheme Annotation Agent} uses prompt-based LLM reasoning to assign an argumentation-scheme set $\mathcal{S}_i$ to each utterance. 
Then, the \textbf{Scoring Agent} performs few-shot evaluation with exemplars $\mathcal{X}_{\text{few}}=\{e_{\text{poor}},e_{\text{general}},e_{\text{good}}
,e_{\text{excellent}}\}$ and outputs \emph{per-scheme} qualitative labels:
\[
\text{Score}_i(S)\in\{\text{poor},\text{general},\text{good},\text{excellent}\}\, \text{for } S\in\mathcal{S}_i.
\]
and $\text{Score}_i(S)=\text{none}$ if $S\notin\mathcal{S}_i$. 
The scoring criteria follow Walton et al.’s guidelines for evaluating argumentation schemes, emphasizing whether the premises are acceptable, the scheme is applied appropriately, and the critical questions are adequately addressed. 
Thus, each label reflects the appropriateness and rhetorical effectiveness of the scheme usage in context.

Finally, each database record is represented as 
$r_i = (u_i, \mathbf{e}_i, \mathcal{S}_i, \text{Score}_i)$, 
constituting the enriched knowledge base $\mathcal{D}=\{r_i\}_{i=1}^N$ that supports retrieval and generation. 
In subsequent stages, $\mathcal{D}$ provides the foundation for stance-aware evidence selection and supports generation modules in reconstructing contextually coherent and strategically grounded arguments.

\subsection{Debate Data Retrieval}

Given an ongoing debate and a target stance $t_{i} \in T$, the retrieval module extracts stance-relevant argumentative context to support the generation of the next utterance. 
It operates through two complementary components: 
a \emph{Logic \& Summarization Agent}, which identifies logic loopholes in the opponent's arguments, 
and a \emph{Keyword Extraction Agent}, which derives salient keywords from the current debate history. 
The extracted keywords are then passed to a basic retriever for coarse-grained evidence retrieval from the knowledge base $\mathcal{D}$, constructed in Section~\ref{sec:DatabaseConstruction}. 
A fine-grained filtering stage then computes semantic similarity between the retrieved candidates and the current debate history to retain the most contextually relevant exemplars.

\smallskip

{\bf Logic \& Summarization Agent.} This component analyzes the accumulated utterances from the opposing side to detect logic loopholes, contradictions, and unsupported assumptions across the entire debate history, rather than at the single-utterance level. 
Inspired by the SymbCoT~\cite{xu2024faithful} paradigm, we design a prompt-based logical decomposition process that guides LLMs to transform the opponent's collective statements into pseudo-first-order predicates and simulate lightweight symbolic reasoning over them.

Formally, given the opponent's utterance sequence defined as $H^{\text{opp}}_{n}=\{u_1^{\text{opp}},\dots,u_t^{\text{opp}}\}$, the agent constructs a set of pseudo–first-order predicates and derives natural-language reasoning chains:
\[
\mathcal{P}_{\text{opp}}=\bigcup_{i=1}^{t} f_{\text{pred}}(u_i^{\text{opp}}), \,
\tau_{\text{opp}} = f_{\text{infer}}(\mathcal{P}_{\text{opp}}),
\]
which summarize the inferred argumentative logic throughout the opponent's discourse.

The first step, \emph{predicate extraction}, converts each utterance into a symbolic form that captures its structure and causal relations.

Once the individual predicates are extracted, the next step reconstructs them into a coherent reasoning chain that represents the opponent's overall argumentative flow.

\begin{tcolorbox}[colback=gray!5!white, colframe=gray!40!black, boxrule=0.3pt, arc=1pt, left=2pt, right=2pt, top=2pt, bottom=2pt]
\small
\textbf{Prompt for $f_{\text{pred}}$:} You are a Symbolic Translator.\\
\textbf{Task:} Read an opponent's debate statement and convert its major argumentative claims into pseudo–first-order predicates. \\
\textbf{Guidelines:}
Identify central claims, causal/temporal relations, and quantitative assertions.
Use concise predicate forms such as:
\textit{Cause(A,B), Supports(A,B), Contrast(A,B), Negate(P)}. 
Exclude rhetorical or filler content, and output one predicate per line. \\
\textbf{Input:} \{[OPPONENT'S  UTTERANCE]\}
\end{tcolorbox}

\begin{tcolorbox}[colback=gray!5!white, colframe=gray!40!black, boxrule=0.3pt, arc=1pt, left=2pt, right=2pt, top=2pt, bottom=2pt]
\small
\textbf{Prompt for $f_{\text{infer}}$:} You are a Symbolic Chain-of-Thought Generator.\\
\textbf{Task:} Take the predicate set extracted by $f_{pred}$ and generate a coherent natural-language reasoning chain representing the opponent's logic.\\
\textbf{Guidelines:}
Interpret causal and contrastive relations among predicates, order them into a logical narrative, and mark where claims support or conflict. 
Maintain neutrality and reconstruct the opponent’s logic, rather than providing a rebuttal. \\
\textbf{Input:} \{[PSEUDO-FIRST-ORDER LOGICAL REPRESENTATION]\}
\end{tcolorbox}

Logical inconsistencies or invalid reasoning patterns in $\tau_{\text{opp}}$ are then identified as
$
\mathcal{L} = f_{\text{logic}}(\tau_{\text{opp}}).
$
where $\mathcal{L}$ denotes a collection of natural-language control signals that provide interpretable guidance for generating targeted rebuttals and reinforcing stance fidelity. 
In this design, only $\mathcal{P}_{\text{opp}}$ corresponds to a pseudo–first-order logical representation, while $\tau_{\text{opp}}$ and $\mathcal{L}$ remain natural-language constructs, enabling symbolic-style interpretability without explicit theorem proving.

To operationalize the three core functions: $f_{\text{pred}}$, $f_{\text{infer}}$, and $f_{\text{logic}}$, we design prompt templates that guide the LLM through predicate extraction, reasoning-chain construction, and logic flaw detection. 
The final stage $f_{\text{logic}}$ verifies the inferred reasoning chain to pinpoint contradictions and unsupported assumptions, producing the interpretable control signal set $\mathcal{L}$ as follows:


\begin{tcolorbox}[colback=gray!5!white, colframe=gray!40!black, boxrule=0.3pt, arc=1pt, left=2pt, right=2pt, top=2pt, bottom=2pt]
\small
\textbf{Prompt for $f_{\text{logic}}$:} You are a Logic Critic / Verifier.\\
\textbf{Task:} Read the reasoning chain and identify logical flaws or inconsistencies, producing a set of natural-language control signals.\\
\textbf{Guidelines:}
Detect contradictions, unsupported assumptions, circular reasoning, or overgeneralization. 
Each flaw should be concise (1–3 sentences) and reference relevant claims.
Do not add new evidence, and only report logical flaws.\\
\textbf{Input:} \{[REASONING CHAIN SET]\}
\end{tcolorbox}

More detailed prompt versions for all three modules 
($f_{\text{pred}}$, $f_{\text{infer}}$, and $f_{\text{logic}}$) 
and an end-to-end example of the three-step process 
are provided in Appendix~\ref{app:logic_case} and Appendix~\ref{app:prompts_d} respectively.

\smallskip

\textbf{Keyword Extraction Agent.} This component operates in parallel with the Logic \& Summarization Agent and focuses on retrieving strategically relevant exemplars from the debate knowledge base $\mathcal{D}$. 
A prompt-based LLM first extracts salient keywords $\mathcal{K}=\{k_1,\dots,k_m\}$ from the partial debate history $H_j=\{u_1,\dots,u_j\}$. 
These keywords are used in a rule-based coarse retrieval stage to filter database records with explicit keyword matches:
\[
\mathcal{E}_{\text{coarse}}=\bigl\{\, r_j \in \mathcal{D}\; \big|\; \texttt{rule\_match}(r_j,\mathcal{K}) \bigr\}.
\]
Subsequently, a fine-grained re-ranking is performed by encoding the current debate history into a single representation $\mathbf{v}_{H}=f_{\text{emb}}(H_j)$ and computing cosine similarity against the embeddings of the coarse candidates $\mathbf{e}_j=f_{\text{emb}}(r_j)$:
\[
\text{sim}(\mathbf{v}_{H},\mathbf{e}_j)=\frac{\mathbf{v}_{H}\cdot \mathbf{e}_j}{\|\mathbf{v}_{H}\|\,\|\mathbf{e}_j\|}, 
\,
\mathcal{E}=\texttt{Top-}k\bigl(\{(r_j,\text{sim}(\mathbf{v}_{H},\mathbf{e}_j)) \mid r_j\in \mathcal{E}_{\text{coarse}}\}\bigr).
\]

Each retrieved chunk $r_i \in \mathcal{R}$ is annotated with a set of argumentation schemes $\mathcal{S}_i$ and corresponding quality labels $\text{Score}_i$ inherited from database construction. 
For aggregation, each label is assigned a numeric value (\emph{poor}=1, \emph{general}=2, \emph{good}=3, \emph{excellent}=4), while schemes absent in a chunk receive $0$. 
For each scheme type $S \in \mathcal{S}$, we then compute its average score across the top-$k$ retrieved candidates:
$
\bar{s}(S) = \frac{1}{k}\sum_{r_i \in \mathcal{E}} \text{Score}_i(S).
$
Schemes with $\bar{s}(S) > 2$ (above the \emph{general} threshold) are retained as high-quality argumentative priors, and the corresponding chunks containing these schemes are selected as exemplars. 
The resulting prior sets $(\mathcal{E}^{\text{prior}}, \mathcal{S}^{\text{prior}})$ jointly form the output of the Keyword Extraction Agent and provide strategy-oriented evidence for subsequent generation.

\subsection{Debate Generation}
Given the retrieved strategic evidence and logical signals from the previous stage, 
the model first generates an initial utterance $\tilde{u}^{(0)}$ conditioned on the debate history $H_n$ and the designated stance $s$: 
\[
\tilde{u}^{(0)} = M(H_n,\,\mathcal{L},\,\mathcal{E}^{\text{prior}},\,\mathcal{S}^{\text{prior}},\,s).
\]
The \emph{Debate Summarization Agent} then analyzes both the accumulated debate history and the newly generated utterance to produce a structured summary covering the overall debate overview, supporting and opposing arguments, and the core points of divergence. 
These structured summaries, together with the candidate utterance, are passed to the judgement Agent.

The \emph{Judgement Agent} evaluates the candidate utterance under three criteria: stance faithfulness, argumentative relevance, and scheme compliance, and then outputs a binary judgement signal $J^{(t)}\!\in\!\{0,1\}$ along with textual feedback $\rho^{(t)}$ explaining any detected violation. 
If the judgement is positive ($J^{(t)}=1$), the utterance is accepted as the final output. 
Otherwise, the \emph{Text Modification Agent} receives both the utterance and the feedback and produces a revised version:
$
\tilde{u}^{(t+1)} = f_{\text{modify}}\big(\tilde{u}^{(t)}, \rho^{(t)}\big).
$
This process repeats until all evaluation criteria are satisfied, and the utterance $u^{*}$ is adopted as the debate output for the turn.

\section{Experiments}
\subsection{Experimental Setup}
\textbf{Dataset.} 
We conduct all experiments on the ORCHID~\cite{zhao2024orchid} debate dataset, which contains high-quality transcripts collected from formal debate tournaments.
Each record contains the debate topic, participating institutions (e.g., Tongji University), speaker identifiers (e.g., Pro-1, Con-2), explicit stance tags (\textit{pro}/\textit{con}), and detailed utterance texts. 
To ensure data consistency, we exclude free-debate or informal discussion sessions during preprocessing.
On average, each debate contains about ten rounds, with each utterance consisting of 1,000--1,500 Chinese characters. 
Since the transcripts originate from standardized competitions with explicit stance and role annotations, ORCHID provides consistent quality and minimizes potential bias across topics and domains.
From the most recent five years of ORCHID, we extract a total of 1,134 debates, of which 1,000 are used to construct the retrieval corpus. For evaluation, we adopt a curated set of 32 debates covering seven representative domains: \emph{Society \& Livelihood}, \emph{Politics \& Governance}, \emph{Economy \& Development}, \emph{Technology \& Future}, \emph{Morality \& Values}, \emph{Education \& Youth}, and \emph{International Relations}, ensuring topical diversity and no overlap with the retrieval corpus.

\textbf{Compared Methods.} 
We compare R-Debater against three baselines. (i) \textbf{LLM:} a direct prompting approach that generates debate utterances without any external retrieval materials or structured planning.  (ii) \textbf{Naive RAG:} a retrieval-augmented baseline that simply concatenates retrieved knowledge to the prompt, without filtering or explicit reasoning control.  (iii) \textbf{Agent4Debate:} a multi-agent coordination framework that assigns distinct roles for analysis, evidence gathering, and rebuttal generation. All baselines are instantiated with \textbf{GPT-4o}~\cite{hurst2024gpt}, \textbf{DeepSeek-V3}~\cite{liu2024deepseek}, and \textbf{Claude-3.7-sonnet}~\cite{Claude3S} under a zero-shot setting, setting the temperature to 0.2 and without fine-tuning, to ensure fairness and reproducibility.

\textbf{Evaluation.} 
We adopt two methods: \textbf{InspireScore}~\cite{wang2025inspiredebate} for utterance-level assessment and \textbf{Debatrix}~\cite{liang2024debatrix} for debate-level evaluation. InspireScore measures utterance quality along three dimensions: Subjective, Logic, and Fact. The \textit{Subjective} dimension captures persuasiveness and rhetorical appeal, including emotional appeal, clarity of the argument, argument arrangement, and topic elevation. \textit{Logic} assesses reasoning validity and stance consistency, ensuring arguments maintain coherent reasoning chains, avoid contradictions, and align with both the assigned position and debate motion. \textit{Fact} evaluates factual grounding and evidence usage, considering correctness, credibility, and contextual relevance of cited information. We report per-dimension results and the aggregated InspireScore averaged across runs. To complement utterance-level analysis, Debatrix evaluates performance by scoring each utterance across three criteria: \textbf{Argument (A)}, \textbf{Source (S)}, and \textbf{Language (L)}, then aggregates them into an \textbf{Overall (O)} score. The final outcome reflects which side achieves superior argument quality, evidence integration, and linguistic fluency, providing a measure of overall debate effectiveness.

\textbf{Experimental Design.}
We design three complementary experiments to evaluate R-Debater from different perspectives: 
 micro-level next-utterance generation, 
 macro-level adversarial debate simulation, 
and internal agent–expert alignment. 
They assess both external performance and internal reliability of the framework.

\textit{Experiment 1: Next-Utterance Generation.}
We assesses the model's ability to generate the next debate utterance, given a truncated dialogue history and stance. 
Each model produces one continuation, which is evaluated using InspireScore in terms of linguistic fluency, argumentative soundness, and stance consistency.

\textit{Experiment 2: Adversarial Debate Simulation.} To assess global debate competence, we conduct simulations in which R-Debater and Agent4Debate engage in multi-turn debates under opposing stances. We initialize each simulation from a balanced stage and allow the debate to proceed until both sides reach a conclusion. The resulting trajectories are evaluated with the Debatrix metric, which measures argument quality, evidence integration, and rhetorical coherence across the debate flow.

\textit{Experiment 3: Expert Alignment Evaluation.} 
To validate interpretability and trustworthiness of the framework, we evaluate alignment between internal agents and human experts. Two components are assessed: \textbf{Scheme Annotation Agent}, which labels argumentation schemes, and \textbf{Scoring Agent}, which rates argumentative quality. For Scheme Annotation Agent, we compute Jaccard Index, Precision, Hamming Loss, Cohen's $\kappa$, and Krippendorff's $\alpha$ to capture surface-level and inter-rater consistency. For Scoring Agent, we examine Pearson, Spearman, and Kendall correlations between automatic and expert scores, reflecting absolute, relative alignment.

\section{Results and Analysis}

\subsection{Single-turn Debate Evaluation}

As shown in Table~\ref{tab:single-turn}, R-Debater consistently surpasses all baselines in single-turn debate performance across different foundation models. 
It achieves the highest scores in all InspireScore dimensions: \textit{Subjective}, \textit{Logic}, and \textit{Fact}, as well as in the overall composite score. 
The most notable gains appear in the logical dimension, where R-Debater approaches a near-perfect score ($\approx$1) across all models, indicating a strong ability to maintain reasoning coherence and stance fidelity. 
Improvements in both the factual and subjective dimensions further demonstrate that integrating retrieval-augmented argumentative memory and multi-agent collaboration enhances both factual grounding and rhetorical persuasiveness, resulting in more convincing and well-structured single-turn arguments.

\subsection{Multi-turn Competitive Debate Evaluation}

We further assess R-Debater in full multi-turn debate scenarios using the Debatrix evaluation framework. 
As shown in Table~\ref{tab:multi-turn}, R-Debater consistently surpasses the Agent4Debate baseline across all evaluation dimensions and foundation models. 
On GPT-4o, it achieves substantial gains in Source Credibility, Language quality, Argument soundness, and Overall performance, while comparable or even greater improvements are observed on DeepSeek-V3. 
These results demonstrate that R-Debater not only sustains its superiority in single-turn reasoning but also exhibits stronger strategic consistency and cross-turn coherence, highlighting the effectiveness of retrieval-augmented argumentative memory and structured multi-agent reasoning in extended debate settings.

\begin{table}[!htbp]
\centering
\small
\caption{Single-turn debate evaluation across models and methods using InspireScore.}
\label{tab:single-turn}
\resizebox{\linewidth}{!}{
\begin{tabular}{l l c c c c}
\hline
\textbf{Base Model} & \textbf{Method} & \textbf{Subjective} & \textbf{Logic} & \textbf{Fact} & \textbf{InspireScore} \\
\hline
\multirow{4}{*}{GPT-4o} 
& LLM & 0.757 & 0.940 & 0.226 & 0.641 \\
& Naive RAG & 0.761 & 0.939 & 0.609 & 0.770 \\
& Agent4Debate & 0.821 & 0.913 & \textbf{0.631} & 0.783 \\
& R-Debater (Ours) & \textbf{0.842} & \textbf{0.997} & 0.627 & \textbf{0.822} \\
\hline
\multirow{4}{*}{DeepSeek-V3} 
& LLM & 0.783 & 0.877 & 0.158 & 0.591 \\
& Naive RAG & 0.732 & 0.927 & 0.582 & 0.747 \\
& Agent4Debate & 0.837 & 0.893 & 0.590 & 0.773 \\
& R-Debater (Ours) & \textbf{0.866} & \textbf{0.997} & \textbf{0.594} & \textbf{0.819} \\
\hline
\multirow{4}{*}{Claude-3.7-sonnet} 
& LLM & 0.780 & 0.971 & 0.190 & 0.647 \\
& Naive RAG & 0.772 & 0.960 & 0.567 & 0.766 \\
& Agent4Debate & 0.765 & 0.917 & 0.618 & 0.767 \\
& R-Debater (Ours) & \textbf{0.789} & \textbf{0.985} & \textbf{0.701} & \textbf{0.830} \\
\hline
\end{tabular}
}
\end{table}

\begin{table}[!htbp]
\centering
\small
\caption{Multi-turn Competitive Debate Evaluation across models and methods using Debatrix.}
\label{tab:multi-turn}
\begin{tabular}{l l c c c c}
\hline
\textbf{Model} & \textbf{Framework} & \multicolumn{4}{c}{\textbf{Debatrix}} \\
\cmidrule(lr){3-6}
& & \textbf{S} & \textbf{L} & \textbf{A} & \textbf{O} \\
\hline
\multirow{2}{*}{GPT-4o} 
& R-Debater (ours) & 1.13 & 1.16 & 1.26 & 1.23 \\
& Agent4Debate & 0.87 & 0.84 & 0.74 & 0.77 \\
\hline
\multirow{2}{*}{Deepseek-V3} 
& R-Debater (ours) & 1.25 & 1.11 & 1.19 & 1.25 \\
& Agent4Debate & 0.75 & 0.89 & 0.81 & 0.75 \\
\hline
\end{tabular}
\end{table}

\subsection{Expert Alignment Evaluation.}

Beyond baseline comparisons, we assess the alignment of R-Debater’s internal agents with expert judgements.

To measure how faithfully the internal modules of R-Debater approximate expert reasoning, we evaluate both the \textbf{Scheme Annotation Agent} and the \textbf{Scoring Agent} against gold-standard expert annotations. 
Table~\ref{tab:agent_eval} presents the agreement metrics for the annotation task and the correlation results for scoring alignment.

\begin{table}[!htbp]
\centering
\caption{Evaluation results of the Scheme Annotation Agent and Scoring Agent.}
\label{tab:agent_eval}
\begin{subtable}[t]{0.48\linewidth}
\centering
\small
\caption{Scheme Annotation Agent vs. expert annotations.}
\label{tab:annotation_agreement}
\begin{tabular}{l c}
\toprule
\textbf{Metric} & \textbf{Value} \\
\midrule
Jaccard Index (avg)      & 0.7366 \\
Precision (avg)          & 0.8225 \\
Hamming Loss (avg)       & 0.1291 \\
Cohen's $\kappa$ (micro) & 0.7229 \\
Cohen's $\kappa$ (macro) & 0.6964 \\
Krippendorff's $\alpha$ (nominal) & 0.7230 \\
\bottomrule
\end{tabular}
\end{subtable}
\hfill
\begin{subtable}[t]{0.48\linewidth}
\small
\centering
\caption{Scoring Agent: correlation with expert scores.}
\label{tab:scoring_agreement}
\begin{tabular}{l c}
\toprule
\textbf{Metric} & \textbf{Value} \\
\midrule
Pearson   & 0.6356 \\
Spearman  & 0.6533 \\
Kendall   & 0.5880 \\
\bottomrule
\end{tabular}
\end{subtable}
\end{table}

\textbf{Annotation Agreement with Experts.}
As shown in Table~\ref{tab:agent_eval} (a), the Scheme Annotation Agent achieves strong agreement with expert annotations across multiple metrics. 
The Jaccard Index and Precision demonstrate that the agent captures the majority of expert-labeled schemes, while low Hamming Loss indicates that misclassifications are rare. 
Furthermore, Cohen’s $\kappa$ (both micro and macro) and Krippendorff’s $\alpha$ reveal substantial inter-rater reliability, confirming that the automated annotation process can approximate expert-level scheme identification with high fidelity. 
These results validate that the retrieval-augmented annotation mechanism produces stable and interpretable scheme assignments.

\textbf{Scoring Alignment with Experts.}
As shown in Table~\ref{tab:agent_eval} (b), the Scoring Agent exhibits high correlation with expert scores (Pearson, Spearman, Kendall), preserving both absolute values and relative rankings. 
This suggests it can serve as a credible proxy for expert evaluation, reducing the reliance on costly human annotation. 
Such alignment also indicates that the scoring module provides a consistent and scalable way to benchmark debate quality, ensuring that R-Debater’s evaluations remain interpretable and trustworthy across large-scale experiments.

Together, these findings indicate that R-Debater’s internal modules align closely with expert reasoning, enhancing the framework’s interpretability and trustworthiness.

\subsection{Case Study}

To illustrate the qualitative advantages of R-Debater, we conducted a case study using data recorded from a real debate experiment in Experiment~2, with the topic ``Should society encourage Generation Z to `rectify' workplace culture?'' R-Debater and Naive RAG were assigned opposite stances under the same historical debate context and tasked with generating the next utterance for their respective sides. Below is an excerpt for context:

\begin{tcolorbox}[colback=gray!5!white, colframe=gray!40!black, boxrule=0.3pt, arc=1pt, left=2pt, right=2pt, top=2pt, bottom=2pt]
\small
\textbf{Opponent (Con):} ``Corporate culture is the foundation of organizational stability, and Gen-Z's questioning of existing systems will only cause internal disorder. Authority and discipline are guarantees of efficiency; we should not overturn the entire system because of a few isolated incidents.''
\end{tcolorbox}

Given this context, the Logic \& Summarization Agent identifies a false dichotomy in the opponent’s reasoning. 
Let $\tau_{\text{opp}}$ denote the set of natural-language reasoning chains extracted from the opponent, the logical control signals are then computed as
\[
\begin{aligned}
\mathcal{L} &= f_{\text{logic}}(\tau_{\text{opp}}) \\[2pt]
            &= \bigl\{\text{``False dichotomy: } 
               (\text{Questioning} \!\rightarrow\! \text{Disorder}) 
               \text{ is invalid.''}\bigr\}.
\end{aligned}
\]
This logical signal $\mathcal{L}$ guides generation of a rebuttal emphasizing the necessity of constructive criticism in organizational evolution.

\begin{table}[h!]
\centering
\small
\caption{Qualitative comparison between R-Debater and Naive RAG on logical reasoning and stance consistency.}
\begin{tabular}{p{0.13\linewidth} p{0.42\linewidth} p{0.36\linewidth}}
\hline
\textbf{Module} & \textbf{R-Debater (Pro)} & \textbf{Naive RAG (Con)} \\
\hline
\textbf{Detected Logical Issue ($\mathcal{L}$)} & 
``Conflates constructive questioning with destructive disorder; assumes all systemic challenges stem from isolated incidents.'' & 
NA. \\
\textbf{Retrieved Evidence ($\mathcal{E}$)} & 
Case: ``Tech company (Meituan) innovation spurred by employee feedback programs that improved both culture and productivity.'' (Argumentation scheme: causal argumentation, score: \textit{good}; exemplification, score: \textit{excellent}) & 
Generic examples of ``corporate hierarchy and discipline'' unrelated to the core argument. \\
\textbf{Generated Rebuttal ($u_{m+1}$)} & 
``Constructive questioning drives organizational adaptation, not disorder. Many successful companies have integrated employee feedback to improve both culture and efficiency, demonstrating that systemic evolution differs from complete overthrow.'' & 
``Encouraging confrontation may undermine teamwork and trust. Rather than opposing authority, fostering open communication would yield more sustainable improvement.'' \\
\hline
\end{tabular}
\label{tab:case_study}
\end{table}

\textbf{Analysis.}
The case highlights R-Debater's ability to reconstruct \textit{argumentative memory} across turns. 
By identifying the false dichotomy between questioning and disorder, R-Debater effectively integrates logical consistency with persuasive coherence. 
It retrieves historically similar cases from $\mathcal{D}$ that show how constructive criticism can foster positive organizational change, thereby directly countering the opponent's premise. 
The generated rebuttal remains focused on the topic while providing factual grounding through concrete examples.

In contrast, Naive RAG fails to engage with the opponent's core reasoning chain, producing surface-level counterpoints that do not address the logical flaw in the original argument. Human evaluation further confirms that R-Debater's outputs exhibit stronger reasoning quality by addressing the opponent's argumentative structure rather than responding to superficial content.

\subsection{Ablation Study}

We conduct a series of ablation studies under the same setting as Experiment~1, using \texttt{gpt-4o-mini} as the base model and InspireScore for evaluation. The results in Table~\ref{tab:ablation_results} demonstrate the complementary roles of all major components in R-Debater.
\begin{table}[!htbp]
\centering
\caption{Ablation results of R-Debater using gpt-4o-mini as the base model. 
``Opt. Module'' / ``L. \& S. Agent'' / ``Argu. Schem.'' remove the Optimization Module, Logic \& Summarization Agent, and Argumentation Augmentation Module, respectively.}
\small
\begin{tabular}{l c c c c}
\hline
\multirow{2}{*}{\textbf{Method}} & \multicolumn{4}{c}{\textbf{InspireScore}} \\
\cline{2-5}
 & \textbf{Subjective} & \textbf{Logic} & \textbf{Fact} & \textbf{InspireScore} \\
\hline
R-debater & 0.841 & 0.954 & 0.700 & 0.831 \\
w/o Opt. Module & 0.773 & 0.917 & 0.640 & 0.776 \\
w/o L. \& S. Agent & 0.735 & 0.875 & 0.672 & 0.761 \\
w/o Argu. Schem. & 0.371 & 0.683 & 0.531 & 0.528 \\
\hline
\end{tabular}
\label{tab:ablation_results}
\end{table}

In all settings, the overall R-Debater framework is preserved, and only one specific component is removed to isolate its contribution. Removing the \emph{Optimization Module} leads to a notable drop in overall InspireScore (0.831$\rightarrow$0.776), primarily driven by declines in the subjective dimension. This indicates that the Optimization Module primarily enhances Subjective performance by improving rhetorical appeal, including clarity, fluency, and emotional resonance, as well as reinforcing the firmness of stance. When the \emph{Logic \& Summarization Agent} is excluded, the decrease (0.831$\rightarrow$0.761) is most pronounced in the logic dimension, confirming that this agent plays a central role in maintaining coherent reasoning chains and reducing logical inconsistencies. The absence of the Argumentation Scheme causes the sharpest degradation (0.831$\rightarrow$0.528), even approaching the naive LLM baseline. Without this mechanism, the model struggles to organize or leverage retrieved argumentative memory, even with the support of the Logic \& Summarization Agent, resulting in fragmented reasoning and factual drift.

Overall, the ablation pattern reveals a clear division of labor within R-Debater: the Optimization Module enhances rhetorical persuasion, the Logic \& Summarization Agent secures reasoning coherence, and the Argumentation Scheme ensures evidence integration. Their synergy underpins the framework’s superior performance in debate generation.

\subsection{Human Evaluation}
Given that the evaluation of debate quality is inherently subjective~\cite{slonim2021autonomous}, we complement automatic metrics with a dedicated human assessment phase. To ensure fairness and reproducibility, all evaluations were conducted under strict randomization and anonymization, eliminating bias from model identity or order effects. This evaluation used the same input set as Experiment~1, and annotators assessed the debate statements generated by the four compared models. Our annotator pool consisted of 20 members from the university debate association, each with demonstrated debating experience and critical reasoning skills.

Following the InspireScore framework, the evaluation covered two categories of dimensions. The \textit{subjective} dimensions included emotional appeal, argument clarity, argument arrangement, and topic relevance, while the \textit{objective} dimensions assessed fact authenticity and logical validity. Each dimension was rated on a 1–10 Likert Scale to capture fine-grained differences in argumentative quality. After completing the dimension-level scoring, annotators selected their overall preferred debate statement among the four candidates, providing an additional holistic measure of preference.

Table~\ref{tab:human_choose} summarizes the human preference rates. Results indicate that R-Debater achieved the highest selection rate ({76.32\%}), substantially outperforming the Agent4Debate (15.79\%),LLM (7.89\%), and NaiveRAG (0\%). This result underscores both the factual grounding and rhetorical persuasiveness of debates generated by R-Debater.

\begin{table}[htbp]
\centering
\small
\caption{Human preference for debate generation models}
\label{tab:human_choose}
\begin{tabular}{lcccc}
\hline
\textbf{Model} & LLM & NaiveRAG & Agent4Debate & R-Debater \\
\hline
\textbf{Choose Rate (\%)} & 7.89 & 0.00 & 15.79 & 76.32 \\
\hline
\end{tabular}
\end{table}

\begin{figure}
    \centering
    \includegraphics[width=0.8\linewidth]{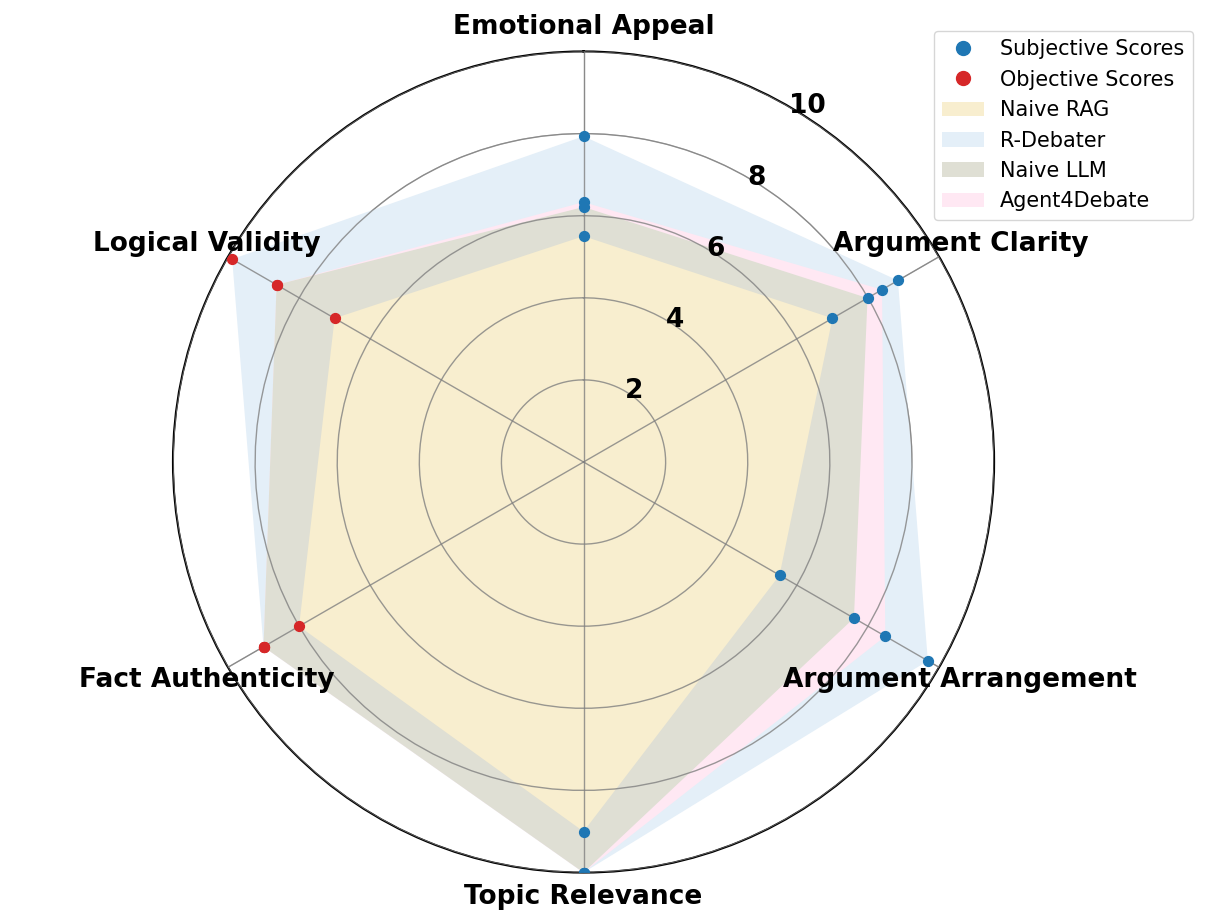}
    \caption{Human evaluation results on the six InspireScore dimensions. Each spoke shows the macro-averaged Likert rating (1-10) for a system; shaded bands indicate 95\% bootstrap CIs.}
    \label{fig:human_inspire}
\end{figure}

Figure~\ref{fig:human_inspire} further reports the macro-averaged Likert ratings across the six InspireScore dimensions, providing a complementary perspective to the overall preference rates.

\section{Limitations}

Although R-Debater shows strong performance, several limitations remain. 
First, the framework inherently depends on pretrained LLMs, which may suffer from outdated knowledge and hallucination issues. 
While retrieval mitigates these problems to some extent, inconsistencies between retrieved evidence and model priors can still compromise factual reliability. 
Second, the retrieval process itself introduces additional latency and potential noise, and the multi-agent architecture further increases token usage and computational overhead, limiting the system’s scalability for real-time or large-scale applications.  
Third, the timeliness and coverage of the debate database play a critical role in overall performance.  
If the repository is not regularly updated or lacks domain diversity, R-Debater may generate arguments that are outdated or topically misaligned. 
Fourth, the current implementation is primarily trained and evaluated on the ORCHID dataset. 
Future work includes extending the framework to cross-lingual and cross-domain settings to assess its generalization capacity, particularly in informal or multilingual debate contexts. 
Finally, metrics such as InspireScore and Debatrix still involve subjective components. 
Although human evaluation improves reliability, it remains small-scale and subjective, underscoring the need for broader and more systematic studies in future assessments.

\section{Conclusion}
We present R-Debater, a retrieval-augmented framework that integrates argumentative memory with structured multi-agent planning to generate coherent, stance-consistent, and evidence-grounded debate arguments across multiple turns. Empirical results on the ORCHID dataset demonstrate that R-Debater outperforms strong LLM-based baselines in factual accuracy, logical coherence, and persuasive quality, validating the effectiveness of reconstructing argumentative memory through retrieval grounding and coordinated agent reasoning. Moving forward, the framework will be extended to broader debate domains and multilingual contexts, with efforts to enhance interpretability and efficiency for real-time interactive use. Together, these advances highlight the potential of retrieval-augmented multi-agent systems to support more transparent, adaptive, and trustworthy debate generation.

\bibliographystyle{ACM-Reference-Format} 
\bibliography{mybibliography}
\appendix
\section*{Appendix}
\addcontentsline{toc}{section}{Appendix}
\appendix
\section{Case Demonstration of the Logic \& Summarization Agent}
\label{app:logic_case}

This appendix presents a concrete, end-to-end example of our three-step pipeline. 
Given an opponent's debate statement, we first extract pseudo-first-order predicates $\mathcal{P}_{\text{opp}}$, then derive a natural-language reasoning chain $\tau_{\text{opp}}$, and finally identify logical flaws $\mathcal{L}$.

\begin{tcolorbox}[
  colback=gray!3!white,
  colframe=black!70!white,
  title={Input: Opponent Debate Statement (Excerpt)},
  fonttitle=\sffamily,
  breakable,
  listing only,
  listing options={breaklines=true, basicstyle=\ttfamily\small}
]
\textbf{Debate Topic:} Has globalization accelerated or slowed in the post-pandemic era?

Con. However, the pandemic has not accelerated globalization; rather, it has disrupted it. The outbreak of COVID-19 nearly halted global economic interactions. Variants spread rapidly in the UK and Germany, while Australia and Malaysia returned to lockdowns. Such disruptions affected people’s lives and forced governments to reassess the impact of globalization.
Data prove this trend: since 2019, global trade dependence has decreased by seven percentage points, whereas it had only decreased by four points over the entire twelve years preceding the pandemic. Foreign direct investment dropped by 38\% due to COVID-19 and remains 25\% below previous levels. Even in 2021, supposedly the ``post-pandemic era,'' indicators continue to decline.
Nations are also adjusting their strategies. President Biden’s American Jobs Plan emphasizes rebuilding domestic supply chains and ensuring Americans ``buy American.'' China’s 14th Five-Year Plan stresses domestic circulation as the core. The pandemic exposed the fragility of over-reliance on global supply chains, prompting governments to turn inward. Moreover, public distrust has grown---polls show that 73\% of respondents believe China should be held responsible for the pandemic, which reduces willingness for international cooperation. Thus, globalization is slowing, not accelerating.

Con. Yet your example of China actually supports our view. China controlled the pandemic well, so its trade grew. But in countries like the UK, Europe, and India---where control failed---borders closed and trade declined. Overall, global trade is still down. The WTO predicts that even by 2022, trade volumes will remain below 2019 levels.
You stress cross-border e-commerce, but this existed before COVID; the pandemic only temporarily amplified it. As for digital supply chains, they are still in their infancy, and there is no long-term evidence they can reverse the slowdown of globalization.
More importantly, the pandemic revealed the necessity of reshoring critical industries. When masks were seized and vaccine production stalled, nations realized that so-called ``risk diversification'' meant leaving eggs in someone else’s basket. In contrast, reshoring ensures essential goods stay secure. That is why industrial reshoring policies---whether in healthcare, steel, or other strategic sectors---are rising worldwide. Globalization, therefore, is not accelerating but rather facing a cautious retreat.
\end{tcolorbox}

\begin{tcolorbox}[
  colback=gray!5!white,
  colframe=black!75!white,
  title={Step 1: Predicate Extraction Result $\;(\mathcal{P}_{\text{opp}})$},
  fonttitle=\sffamily,
  breakable,
  listing only,
  listing options={breaklines=true, basicstyle=\ttfamily\small}
]
PredicateSet = \{
  Disrupts(Pandemic, Globalization),
  NearlyHalts(Pandemic, GlobalEconomicInteractions),
  SpreadVariants(\{UK, Germany\}),
  Lockdowns(\{Australia, Malaysia\}),
  Decrease(TradeDependence, since2019, 7pp),
  Decrease(TradeDependence, pre2019\_12y, 4pp),
  Decrease(FDI, Pandemic, 38\%),
  BelowBaseline(FDI, Current, 25\%),
  ContinuesDecline(Indicators, 2021),
  Emphasizes(AmericanJobsPlan, DomesticSupplyChains \& BuyAmerican),
  Emphasizes(China14FYP, DomesticCirculation),
  Reveals(Pandemic, Fragility(GlobalSupplyChains)),
  Increases(PublicDistrust, Pandemic),
  Blames(Public, China, 73\%),
  Concludes(Overall, GlobalizationSlowing),
  Notes(ChinaTrade, GrewUnderGoodControl),
  Notes(\{UK, Europe, India\}, BordersClosed \& TradeDeclined),
  Predicts(WTO, TradeVolumeBelow(2022, 2019)),
  Claims(CrossBorderEcommerce, PreExisted \& TemporarilyAmplified),
  Claims(DigitalSupplyChains, EarlyStage \& InsufficientEvidenceLongTermReversal),
  Advocates(Reshoring, CriticalIndustries),
  MotivatedBy(Reshoring, SecurityOfEssentialGoods),
  Observes(ReshoringPoliciesRising, \{Healthcare, Steel, StrategicSectors\})
\}
\end{tcolorbox}

\begin{tcolorbox}[
  colback=gray!5!white,
  colframe=black!75!white,
  title={Step 2: Natural-Language Reasoning Chain $\;(\tau_{\text{opp}})$},
  fonttitle=\sffamily,
  breakable,
  listing only,
  listing options={breaklines=true, basicstyle=\ttfamily\small}
]
ReasoningChain = """
The opponent argues that the pandemic disrupted globalization, as evidenced by halted interactions, renewed lockdowns, and the spread of variants across multiple countries. 
They further claim that quantitative indicators confirm a slowdown: global trade dependence fell by 7 percentage points since 2019 (compared with only 4 points in the prior twelve years), FDI dropped by 38\% and remains 25\% below previous levels, and key indicators continued to decline in 2021.
Policy responses are presented as consistent with de-globalization: the American Jobs Plan promotes domestic supply chains and Buy American, while China's 14th Five-Year Plan stresses domestic circulation. 
The pandemic is said to reveal global supply-chain fragility, increasing public distrust (73\% blaming China), which purportedly reduces willingness for international cooperation. 
Even if some economies (e.g., China) sustained or grew trade under effective control, the opponent maintains that overall global trade remains below 2019 levels, citing a WTO prediction that 2022 trade volume would still be lower than in 2019.
They contend that cross-border e-commerce predated COVID and was only temporarily amplified, and that digital supply chains are nascent without long-term evidence of reversing the slowdown. 
From the security perspective, they argue that shortages and seizures during the pandemic justify reshoring in critical sectors; the observed rise of reshoring policies is then taken as a sign of a broader, cautious retreat from globalization. 
Hence, the conclusion is that globalization is slowing rather than accelerating.
"""
\end{tcolorbox}

\begin{tcolorbox}[
  colback=gray!5!white,
  colframe=black!75!white,
  title={Step 3: Detected Logical Flaws $\;(\mathcal{L})$},
  fonttitle=\sffamily,
  breakable,
  listing only,
  listing options={breaklines=true, basicstyle=\ttfamily\small}
]
LogicalFlaws = \{
  "Temporal Scope Ambiguity: The argument mixes short-term shock (lockdowns, 2020--2021 contractions) with a long-term trend claim ('globalization is slowing') without clearly separating transient effects from persistent structural change.",
  "Selective Quantification: Declines in FDI and trade dependence are emphasized, but the chain does not examine heterogeneity across regions/sectors (e.g., China trade growth) and whether rebounds or offsetting channels (services, digital trade) exist.",
  "Causal Leap: Public distrust (73\% blaming China) is treated as directly reducing international cooperation, but the mechanism is not specified (policy channels, durations, or countervailing incentives).",
  "Equivocation Between Reshoring and Deglobalization: Reshoring in critical sectors is taken as general evidence of global retreat, yet partial risk rebalancing can co-exist with continued cross-border integration elsewhere.",
  "Status-Quo Comparison: The claim that 2022 trade volumes remain below 2019 is used to infer a slowdown, but no baseline for expected recovery, cyclical adjustment, or substitution effects is provided.",
  "Persistence Assumption: The narrative assumes that amplified cross-border e-commerce and digital supply-chain adoption are temporary, without engaging with data that could indicate durable post-shock adoption."
\}
\end{tcolorbox}

\section{Prompts for Scheme Annotation Agent}

The following prompts describe various argumentative schemes that the Scheme Annotation Agent utilizes during debate annotation and analysis. Each scheme is defined with its method, definition, scenario, example, analytical process, and sample output.

\begin{tcolorbox}[
  colback=gray!5!white,
  colframe=black!75!white,
  title={Instruction \& Input Template for the Scheme Annotation Agent},
  fonttitle=\sffamily,
  breakable,
  listing only,
  listing options={breaklines=true, basicstyle=\ttfamily\small}
]
You are the \textbf{Scheme Annotation Agent}. Your task is to read a single debate utterance and, using the scheme definitions below, label all argumentation schemes that appear in the utterance (multi-label is allowed).
\\

Allowed scheme names (use exactly these as JSON keys) and their descriptions:\\
\{[SCHEME LIST]\}

Argumentation Scheme Details: \\
\{[SCHEME DESCRIPTIONS]\}
\\

What to output: \\
Return a single JSON object where each present scheme is a key and the value is an object with a concise English explanation under the field \texttt{"reason"} that cites the evidence from the utterance (no more than 40 words per reason). Do not invent new scheme names.
\\

Formatting rules:\\
1) Output only raw JSON (no extra text, no Markdown fences, no trailing commas). \\
2) If no scheme is found, output \texttt{\{\}}. \\
3) Reasons must be in English and refer to the utterance content. \\
4) If multiple schemes appear, include all of them as separate keys.
\\

Expected JSON format:\\
\{ \\
\ \ \ "Example-Based Argumentation": \{"reason": "Points to a specific case to generalize the conclusion."\}, \\
\ \ \ "Causal Argumentation": \{"reason": "Claims X leads to Y via an explicit cause-effect link."\} \\
\}
\\

Utterance to annotate: \\
\{[USER INPUT]\}
\\

\end{tcolorbox}

The following argumentation schemes were selected and used in our experiments.

1. Example-Based Argumentation


\begin{tcolorbox}[
  colback=gray!5!white,
  colframe=black!75!white,
  title={Example-Based Argumentation},
  fonttitle=\sffamily,
  breakable,
  listing only,
  listing options={breaklines=true, basicstyle=\ttfamily\small}
]
\textbf{Definition}: This scheme supports a general conclusion through specific cases. 
It assumes that the attributes of an individual or event are representative and can be generalized to similar situations. 
Its validity depends on whether the case truly reflects the broader situation. 
It also allows questioning the representativeness of the opponent’s case. \\
Scenario: Common in policy evaluation or ethical debates, where a debater cites specific instances to support claims. 
The opponent may challenge the applicability in different contexts. \\

\textbf{Example:} ``Although your city-level data shows the policy is effective, it does not cover rural areas, hence it cannot prove nationwide applicability.'' \\
\textbf{Analytical Process:} Based on inductive reasoning, assuming case universality, followed by examining case limitations to challenge the logical rigor of the argument. \\
\textbf{Sample Output:} Argumentative Technique: Example-Based Argumentation. 
Logical Basis: Reveals that case representativeness does not hold, invalidating generalization.
\end{tcolorbox}

2. Value-Based Argumentation

\begin{tcolorbox}[
  colback=gray!5!white,
  colframe=black!75!white,
  title={Value-Based Argumentation},
  fonttitle=\sffamily,
  breakable,
  listing only,
  listing options={breaklines=true, basicstyle=\ttfamily\small}
]
\textbf{Definition:} This scheme relies on ethical or moral core values to assess whether a policy, behavior, or opinion conforms to a particular value system. Its logic centers on identifying value standards and evaluating legitimacy accordingly. \\
\textbf{Scenario:} In debates involving value conflicts, debaters argue from differing moral standards and value orientations, weighing priorities to build reasoning. \\

\textbf{Example:} ``Although economic growth is important, environmental protection must take precedence as the foundation of sustainable development for long-term welfare.'' \\
\textbf{Analytical Process:} First, clarify the core value(s); then link competing values with clear priority ordering to argue for or against compliance. \\
\textbf{Sample Output:} Argumentative Technique: Value-Based Argumentation. Logical Basis: Constructs legitimacy reasoning from core values and their priority.
\end{tcolorbox}

3. Expert Opinion Argumentation

\begin{tcolorbox}[
  colback=gray!5!white,
  colframe=black!75!white,
  title={Expert Opinion Argumentation},
  fonttitle=\sffamily,
  breakable,
  listing only,
  listing options={breaklines=true, basicstyle=\ttfamily\small}
]
\textbf{Definition:} This scheme uses expert statements, professional knowledge, and recognized authority to validate premises or conclusions, often by clarifying key concepts to strengthen credibility. \\
\textbf{Scenario:} In academic, technical, or professional debates, expert viewpoints bolster trust and reliability. \\

\textbf{Example:} ``As a renowned psychologist noted, anger can trigger a sense of justice and thus should not be dismissed as merely negative.'' \\
\textbf{Analytical Process:} Introduce a qualified expert, establish domain relevance and independence, clarify key terms, and connect the expert claim to the conclusion with warranted backing. \\
\textbf{Sample Output:} Argumentative Technique: Expert Opinion Argumentation. Logical Basis: Strengthens reasoning through authoritative, domain-relevant support.
\end{tcolorbox}

4. Positive Consequence Argumentation

\begin{tcolorbox}[
  colback=gray!5!white,
  colframe=black!75!white,
  title={Positive Consequence Argumentation},
  fonttitle=\sffamily,
  breakable,
  listing only,
  listing options={breaklines=true, basicstyle=\ttfamily\small}
]
\textbf{Definition:} This scheme justifies an action or policy by emphasizing its beneficial outcomes, making the argument intuitive and persuasive via demonstrated gains. \\
\textbf{Scenario:} In law reform or policy evaluation, debaters highlight historical progress or present benefits to justify adoption or continuation. \\

\textbf{Example:} ``The modern legal system ensures power supervision and promotes fairness and justice.'' \\
\textbf{Analytical Process:} Marshal empirical evidence and historical comparisons to show a credible pathway from action to desirable effects; address feasibility and scope. \\
\textbf{Sample Output:} Argumentative Technique: Positive Consequence Argumentation. Logical Basis: Demonstrates rationality through a credible positive causal chain.
\end{tcolorbox}

5. Causal Argumentation

\begin{tcolorbox}[
  colback=gray!5!white,
  colframe=black!75!white,
  title={Causal Argumentation},
  fonttitle=\sffamily,
  breakable,
  listing only,
  listing options={breaklines=true, basicstyle=\ttfamily\small}
]
\textbf{Definition:} Establishes a clear causal relation, arguing that a premise inevitably or with high likelihood leads to a result; focuses on the coherence and plausibility of the causal chain. \\
\textbf{Scenario:} In technology, policy, or ethics debates, causal links are frequently contested. \\

\textbf{Example:} ``Without regulation, technological advances will inevitably cause privacy violations, harming society.'' \\
\textbf{Analytical Process:} Specify mechanisms, conditions, and intermediaries; test for alternative causes and hidden assumptions; evaluate necessity/sufficiency. \\
\textbf{Sample Output:} Argumentative Technique: Causal Argumentation. Logical Basis: Validates rigor through explicitly articulated and tested causal links.
\end{tcolorbox}

6. Negative Consequence Argumentation

\begin{tcolorbox}[
  colback=gray!5!white,
  colframe=black!75!white,
  title={Negative Consequence Argumentation},
  fonttitle=\sffamily,
  breakable,
  listing only,
  listing options={breaklines=true, basicstyle=\ttfamily\small}
]
\textbf{Definition:} Refutes an opponent’s standpoint by deriving unacceptable or extreme negative consequences from it, thereby challenging its plausibility or desirability. \\
\textbf{Scenario:} Used in risk management or ethical debates to highlight dangers of adopting the opponent’s premise or policy. \\

\textbf{Example:} ``If we accept the idea that ‘the harder, the better’, then even dangerous activities like rooftop jumps would be justified, which is absurd.'' \\
\textbf{Analytical Process:} Assume the opponent’s premise, project it along a plausible path to extreme or harmful outcomes, and expose contradictions or rule violations. \\
\textbf{Sample Output:} Argumentative Technique: Negative Consequence Argumentation. Logical Basis: Reveals absurd or harmful outcomes entailed by the opponent’s logic.
\end{tcolorbox}

7. Analogical Argumentation

\begin{tcolorbox}[
  colback=gray!5!white,
  colframe=black!75!white,
  title={Analogical Argumentation},
  fonttitle=\sffamily,
  breakable,
  listing only,
  listing options={breaklines=true, basicstyle=\ttfamily\small}
]
\textbf{Definition:} Supports a conclusion by appealing to similarity between two situations with comparable logical structure, contingent on verified shared properties. \\
\textbf{Scenario:} In public policy or social reform debates, speakers cite cross-domain or cross-country cases to suggest transferable solutions. \\

\textbf{Example:} ``Just as country X achieved progress after adopting this policy, it can also promote development here with appropriate adjustments.'' \\
\textbf{Analytical Process:} Identify the base and target cases, establish relevant commonalities, check disanalogies, and infer a parallel conclusion with stated limits. \\
\textbf{Sample Output:} Argumentative Technique: Analogical Argumentation. Logical Basis: Validates universality and rationality via warranted analogy.
\end{tcolorbox}

\section{Prompts for Scoring Agent}

The following prompts describe how the Scoring Agent evaluates the quality of scheme usage in debate utterances. Each dimension of evaluation is defined with its criteria, scenario, example, analytical process, and sample output.

\begin{tcolorbox}[
  colback=gray!5!white,
  colframe=black!75!white,
  title={Instruction \& Input Template for the Scoring Agent},
  fonttitle=\sffamily,
  breakable,
  listing only,
  listing options={breaklines=true, basicstyle=\ttfamily\small}
]
You are the \textbf{Scoring Agent}. Your task is to read a single debate utterance and, for each annotated scheme, assign a quality label from the set:
\{poor, general, good, excellent\}. If the scheme is not used, output \texttt{none}.
\\

Evaluation focuses on four key dimensions:\\
1) Acceptability \\
2) Inference \& Defeasibility \\
3) Relevance \\
4) Robustness \& Rhetorical Clarity \\
\\
What to output: \\
Return a single JSON object where each scheme is a key and the value is an object with dimension-level scores. Scores must be one of \{poor, general, good, excellent, none\}.
\\

Formatting rules:\\
1) Output only raw JSON (no extra text, no Markdown fences, no trailing commas). \\
2) If no scheme is used, output \{\}. \\
3) Reasons must be in English and cite the utterance content concisely. \\
4) Each scheme is evaluated independently with four dimensions.
\\

Expected JSON format:\\
\{ \\
\ \ \ "Causal Argumentation": \{"Acceptability": "good", "Inference": "excellent", "Relevance": "good", "Defeasibility": "general"\}, \\
\ \ \ "Expert Opinion Argumentation": \{"Acceptability": "excellent", "Inference": "good", "Relevance": "good", "Defeasibility": "good"\} \\
\}
\\

Utterance to evaluate: \\
\{[SCHEME ANNOTATION AGENT OUTPUT]\}
\\
\end{tcolorbox}


The following are the definitions of dimensions:

1. Acceptability

\begin{tcolorbox}[
  colback=gray!5!white,
  colframe=black!75!white,
  title={Acceptability},
  fonttitle=\sffamily,
  breakable,
  listing only,
  listing options={breaklines=true, basicstyle=\ttfamily\small}
]
\textbf{Definition:} Judges whether the premises and assumptions are credible and contextually defensible. \\
\textbf{Scenario:} Used when a debater relies on explicit or implicit premises to support reasoning. \\

\textbf{Example:} ``Government data shows rising pollution levels, so regulation is urgent.'' \\
\textbf{Analytical Process:} Examine the trustworthiness of data and assumptions; weak or anecdotal evidence reduces acceptability. \\
\textbf{Sample Output:} Acceptability: good. Premises are factual, plausible, and contextually relevant.
\end{tcolorbox}

2. Inference \& Defeasibility

\begin{tcolorbox}[
  colback=gray!5!white,
  colframe=black!75!white,
  title={Inference \& Defeasibility},
  fonttitle=\sffamily,
  breakable,
  listing only,
  listing options={breaklines=true, basicstyle=\ttfamily\small}
]
\textbf{Definition:} Evaluates whether the reasoning follows the canonical pattern of the scheme and respects its defeasible nature. \\
\textbf{Scenario:} In causal or expert opinion schemes, check if the reasoning anticipates critical questions. \\

\textbf{Example:} ``Without regulation, new technology will inevitably cause privacy risks.'' \\
\textbf{Analytical Process:} Test whether reasoning is overly deductive, ignores exceptions, or misuses the scheme. \\
\textbf{Sample Output:} Inference: good. Reasoning is consistent with the scheme and addresses counterpoints.
\end{tcolorbox}

3. Relevance

\begin{tcolorbox}[
  colback=gray!5!white,
  colframe=black!75!white,
  title={Relevance},
  fonttitle=\sffamily,
  breakable,
  listing only,
  listing options={breaklines=true, basicstyle=\ttfamily\small}
]
\textbf{Definition:} Assesses whether the chosen scheme truly fits the reasoning pattern and conclusion. \\
\textbf{Scenario:} In analogical or value-based schemes, check if the cited case or value is truly pertinent. \\

\textbf{Example:} ``Like country X, we should adopt universal healthcare policies.'' \\
\textbf{Analytical Process:} Identify if reasoning aligns with the intended scheme or introduces irrelevant elements. \\
\textbf{Sample Output:} Relevance: excellent. The argument closely aligns with the scheme's purpose.
\end{tcolorbox}

4. Robustness \& Rhetorical Clarity

\begin{tcolorbox}[
  colback=gray!5!white,
  colframe=black!75!white,
  title={Robustness \& Rhetorical Clarity},
  fonttitle=\sffamily,
  breakable,
  listing only,
  listing options={breaklines=true, basicstyle=\ttfamily\small}
]
\textbf{Definition:} Measures resilience to objections and clarity of delivery. \\
\textbf{Scenario:} When arguments risk collapse under counterexamples or unclear phrasing. \\

\textbf{Example:} ``If we follow your logic, absurd actions like reckless stunts would be justified.'' \\
\textbf{Analytical Process:} Evaluate ability to anticipate counterarguments and maintain persuasive coherence. \\
\textbf{Sample Output:} Robustness: general. Argument is somewhat clear but vulnerable to objections.
\end{tcolorbox}


\begin{tcolorbox}[
  colback=gray!5!white,
  colframe=black!75!white,
  title={Few-Shot Exemplars},
  fonttitle=\sffamily,
  breakable,
  listing only,
  listing options={breaklines=true, basicstyle=\ttfamily\small}
]
\textbf{Good Example 1:} \\
Utterance: ``Because decades of industrial pollution have led to elevated disease rates in neighboring towns, we must adopt stricter environmental regulations now.'' \\
Evaluation: Acceptability: good; Inference: good; Relevance: excellent; Robustness: good. \\

\textbf{Good Example 2:} \\
Utterance: ``Many nations with universal healthcare report better outcomes; adopting a similar system is likely beneficial here.'' \\
Evaluation: Acceptability: good; Inference: good; Relevance: excellent; Robustness: good. \\

\textbf{Bad Example 1:} \\
Utterance: ``I saw two sick people near the factory yesterday, so the factory must be causing all illnesses in this region.'' \\
Evaluation: Acceptability: general; Inference: poor; Relevance: general; Robustness: poor. \\

\textbf{Bad Example 2:} \\
Utterance: ``Because a famous scientist once said this chemical is dangerous, we must ban it immediately.'' \\
Evaluation: Acceptability: general; Inference: poor; Relevance: general; Robustness: poor.
\end{tcolorbox}


\section{Prompts for Methods in Logic \& Summarization Agent}
\label{app:prompts_d}

To ensure interpretability and reproducibility, we provide the prompt templates used for the three core reasoning components in the Logic \& Summarization Agent. 
Each method plays a distinct role in transforming an opponent's argument into structured, symbolic, and interpretable reasoning signals. 
While the detailed prompts are listed below, the same logic pipeline is consistently applied across all debate scenarios.

\begin{tcolorbox}[
  colback=gray!5!white,
  colframe=black!75!white,
  title={Predicate Extraction (\(f_{\text{pred}}\))},
  fonttitle=\bfseries,
  breakable,
  listing only,
  listing options={breaklines=true, basicstyle=\ttfamily\small}
]

\small
Role: You are a Symbolic Translator. \\
Task: Read an opponent's debate statement and convert its major argumentative claims into pseudo-first-order logical predicates. \\

Guidelines:
- Identify central claims, causal/temporal relations, and quantitative assertions. 
- Use compact predicate forms (Cause(A,B), Necessary(A,B), Supports(A,B), Increase(X,Value), Decrease(X,Value), Contrast(A,B), Negate(P)).
- If numbers, percentages, or time periods appear, include them. 
- Use consistent entity names 
- Avoid rhetorical/filler content; only capture logical claims.
- Output only predicates (one per line), not natural sentences. \\

Prompt format:
Here is the opponent's statement:
\{[STATEMENT TEXT]\}

Please output a set of pseudo-first-order predicates capturing the core argumentative logic.
\end{tcolorbox}


\begin{tcolorbox}[
  colback=gray!5!white,
  colframe=black!75!white,
  title={Reasoning Chain Construction (\(f_{\text{infer}}\))},
  fonttitle=\bfseries,
  breakable,
  listing only,
  listing options={breaklines=true, basicstyle=\ttfamily\small}
]
\small
Role: You are the Symbolic Chain-of-Thought Generator. \\
Task: Take the predicate set extracted by $f_{pred}$ and generate a natural-language reasoning chain (\(\tau_{\mathrm{opp}}\)). \\

Guidelines:
- Read the predicate list and interpret relations (cause, support, contrast, contradiction).
- Order them into a coherent reasoning narrative.
- Explicitly mark where claims support or conflict.
- Maintain neutrality (reconstruct opponent's logic, not rebuttal).
- Do not add new evidence; only use given predicates. \\

Prompt format:
Here is the predicate set from the opponent:
{ p1, p2, p3, … }

Please produce a step-by-step reasoning narrative (\(\tau_{\mathrm{opp}}\)) in natural language that links these predicates, indicating support, causation, and possible contradictions.
\end{tcolorbox}


\begin{tcolorbox}[
  colback=gray!5!white,
  colframe=black!75!white,
  title={Logical Flaw Detection (\(f_{\text{logic}}\))},
  fonttitle=\bfseries,
  breakable,
  listing only,
  listing options={breaklines=true, basicstyle=\ttfamily\small}
]
\small
Role: You are the Logic Critic / Verifier. \\
Task: Read the reasoning chain (\(\tau_{\mathrm{opp}}\)) generated by $f_{infer}$ and identify logical flaws or inconsistencies in nature language($\mathcal{L}$). \\

Guidelines:
- Detect contradictions, unsupported assumptions, circular reasoning, overgeneralization, temporal mismatches, or missing links. 
- Each flaw is concise (1--3 sentences), in natural language. 
- Reference relevant predicates or claims explicitly. 
- Do not counterargue with new evidence; only report logical flaws. 
- Output a list of flaw descriptions. \\

Prompt format:
Here is the reasoning chain from the opponent (\(\tau_{\mathrm{opp}}\)):
<(\(\tau_{\mathrm{opp}}\)) text>

Please identify all logical flaws or inconsistencies in the chain, and for each, output a natural-language description referencing the relevant claims.

\end{tcolorbox}


\clearpage
\onecolumn
\begingroup
\setlength{\parindent}{0pt}
\renewcommand{\arraystretch}{1.25}
\setlength{\tabcolsep}{6pt}
\section{Example of Human Evaluation Form}
\label{appendix:human}
\textbf{Debate Topic:} Has globalization accelerated or slowed in the post-pandemic era?\\[1.0em]

\textbf{Background.} You are invited to participate in a study on automatic debate generation. Each task provides (i) an incomplete debate context and (ii) four candidate debate statements generated by different systems. Your role is to evaluate these candidate statements according to the given criteria. All responses are anonymous and will only be used for research purposes.\\[0.8em]

\textbf{Debate Context}\\[0.2em]
\textbf{Pro.} The post-pandemic era refers to the period when the pandemic has largely subsided, with only occasional small outbreaks. Globalization refers to the strengthening of cross-border flows and the deepening of interdependence among nations. We argue that the post-pandemic era has accelerated globalization for three reasons.

First, economic demand. The economic pressures amplified by the pandemic have forced companies to rely even more on the global division of labor and supply chains. For example, labor costs in the U.S. pharmaceutical industry are five times higher than in India, and American manufacturing costs are more than 30\% higher than in China. Withdrawing from globalization would dramatically increase costs. According to the Institute of International Finance, global corporate debt rose by \$5.4 trillion in 2020, and default rates reached their highest since the financial crisis. Under such debt pressure, rejoining global supply chains is inevitable. CPB monitoring data show that by November 2020, global trade had already returned to pre-pandemic levels, and in March 2021, international trade grew by 2.3\%. This rapid recovery is solid evidence that globalization has accelerated.\\[0.6em]

\textbf{Con.} However, the pandemic has not propelled globalization forward; rather, it has disrupted it. The outbreak of COVID-19 nearly halted global economic interactions. Variants spread rapidly in the UK and Germany, while Australia and Malaysia returned to lockdowns. Such disruptions affected people’s lives and forced governments to reassess the impact of globalization.

Data prove this trend: since 2019, global trade dependence has decreased by seven percentage points, whereas it had only decreased by four points over the entire twelve years preceding the pandemic. Foreign direct investment dropped by 38\% due to COVID-19 and remains 25\% below previous levels. Even in 2021, supposedly the “post-pandemic era,” indicators continue to decline.

Nations are also adjusting their strategies. President Biden’s American Jobs Plan emphasizes rebuilding domestic supply chains and ensuring Americans “buy American.” China’s 14th Five-Year Plan stresses domestic circulation as the core. The pandemic exposed the fragility of over-reliance on global supply chains, prompting governments to turn inward. Moreover, public distrust has grown—polls show that 73\% of respondents believe China should be held responsible for the pandemic, which reduces willingness for international cooperation. Thus, globalization is slowing, not accelerating.\\[0.6em]

\textbf{Pro.} The opposition emphasizes reshoring, but this cannot truly mitigate risk. If a domestic outbreak shuts down production, a purely national supply chain still collapses. Real “not putting all eggs in one basket” means having multiple global suppliers, enabled by digitalized supply chains. The pandemic accelerated digital globalization: big data and IoT allow firms to monitor supply and demand shifts, reduce costs, and optimize distribution. A U.S. Congressional report estimates digitalized supply chains can cut unit costs by 30--40\%.

Cross-border e-commerce has also surged, deepening global ties. Meanwhile, reshoring imposes massive costs: rebuilding domestic factories and hiring domestic workers makes products more expensive, reducing competitiveness at a time when both companies and consumers are financially strained. Evidence supports this: once China controlled the pandemic in 2021, outbound investment rose 27.1\% compared to 2019. This proves that once risks are managed, firms quickly return to global markets. Thus, in terms of risk, cost, and post-pandemic data, globalization is accelerating.\\[0.6em]

\textbf{Con.} Yet your example of China actually supports our view. China controlled the pandemic well, so its trade grew. But in countries like the UK, Europe, and India—where control failed—borders closed and trade declined. Overall, global trade is still down. The WTO predicts that even by 2022, trade volumes will remain below 2019 levels.

You stress cross-border e-commerce, but this existed before COVID; the pandemic only temporarily amplified it. As for digital supply chains, they are still in their infancy, and there is no long-term evidence they can reverse the slowdown of globalization.

More importantly, the pandemic revealed the necessity of reshoring critical industries. When masks were seized and vaccine production stalled, nations realized that so-called “risk diversification” meant leaving eggs in someone else’s basket. In contrast, reshoring ensures essential goods stay secure. That is why industrial reshoring policies—whether in healthcare, steel, or other strategic sectors—are rising worldwide. Globalization, therefore, is not accelerating but rather facing a cautious retreat.\\[0.8em]

\textbf{Candidate 1}\\
In the post-pandemic era, globalization has not slowed but accelerated due to multiple factors. First, the core driver of economic globalization—global division of labor and supply chain optimization—has become even more important after the pandemic. Companies face enormous cost pressures, and data show that in 2020 global corporate debt increased by \$5.4 trillion, with default rates at their highest since the financial crisis. This forces firms to rejoin global supply chains in the post-pandemic era to reduce costs. International trade recovered in 2021 and continues to grow. Second, the pandemic accelerated the digital economy, with digitalized supply chains tightening global networks. Cross-border e-commerce grew 53\% in Q2 2020, and in the first half of 2021, China’s cross-border e-commerce imports and exports rose 28.6\%. Digital globalization significantly reduced supply chain costs. Third, global risks such as pandemics and climate change demand closer cooperation among nations. The COVAX program’s successful implementation shows countries increasingly rely on globalization to address shared risks. Thus, economic, technological, and governance needs in the post-pandemic era jointly drive globalization forward.\\[0.6em]

\textbf{Candidate 2}\\
The post-pandemic era refers to the period when the pandemic has largely subsided. Globalization means the strengthening of cross-border flows and deepening interdependence among nations. We argue that the post-pandemic era accelerates globalization for three reasons. First, economic demand: pandemic-driven pressures force firms to remain globalized to reduce costs. Second, the digital economy has been accelerated, with digitalized supply chains tightening global networks and fostering cross-border e-commerce. Third, the need to cope with global risks makes countries part of a shared risk community, where global governance systems become key. Thus, globalization accelerates in the post-pandemic era.\\[0.6em]

\textbf{Candidate 3}\\
The post-pandemic era refers to the period when the pandemic has largely subsided with occasional outbreaks. Globalization means strengthening cross-border flows and interdependence. We argue that globalization has accelerated for three reasons. First, the core driver of globalization is cost reduction through global division of labor. Pandemic-driven cost pressures push firms to stay global. For example, U.S. pharmaceutical labor costs are five times those in India; withdrawing would sharply increase costs. By November 2020, global trade had recovered to pre-pandemic levels, and in March 2021, international trade grew 2.3\%, showing acceleration. Second, the pandemic accelerated the digital economy: digital globalization fosters efficient global data flows, with big data reducing production costs. Studies show digital supply chains can lower unit costs by 30--40\%. Cross-border e-commerce grew 53\% in 2020, and in 2021, China’s cross-border e-commerce trade rose 28.6\%, showing digital economy support for globalization. Third, shared risk management increases interdependence, as seen in WHO’s COVAX program, which highlights global cooperation. Thus, globalization is accelerating in the post-pandemic era.\\[0.6em]

\textbf{Candidate 4}\\
Thank you, Chair. The opposition tried to argue that pandemic-era trade declines and reshoring prove globalization slowed. But this reasoning is logically flawed. First, their evidence cherry-picks short-term pandemic data, ignoring long-term recovery. By March 2021, global trade grew 2.3\%, showing resilience. Second, the digital economy is not a temporary trend: the pandemic catalyzed its adoption, with cross-border e-commerce and IoT integration driving deeper globalization. Third, reshoring examples are isolated and economically unsustainable. Japan’s few reshored firms pale in comparison to the inflows of new foreign investment in China. Ultimately, globalization is driven by efficiency and cost optimization, which reshoring cannot overturn. Ultimately, global risks foster international cooperation, as evident in the sharing of vaccines through COVAX. Thus, globalization in the post-pandemic era is accelerating, not retreating.\\[1.0em]

\textbf{Evaluation Criteria (1--10 each):} \textit{Emotional Appeal, Argument Clarity, Argument Arrangement, Topic Relevance, Fact Authenticity, Logical Validity}.\\[0.6em]

\clearpage
{\large \textbf{Scoring Table: Candidate 1}}\\[0.1em]

\begin{tabular}{lcccccccccc}
\toprule
\textbf{Dimension} & 1 & 2 & 3 & 4 & 5 & 6 & 7 & 8 & 9 & 10 \\
\midrule
Emotional Appeal     & $\square$ & $\square$ & $\square$ & $\square$ & $\square$ & $\square$ & $\square$ & $\square$ & $\square$ & $\square$ \\
Argument Clarity     & $\square$ & $\square$ & $\square$ & $\square$ & $\square$ & $\square$ & $\square$ & $\square$ & $\square$ & $\square$ \\
Argument Arrangement & $\square$ & $\square$ & $\square$ & $\square$ & $\square$ & $\square$ & $\square$ & $\square$ & $\square$ & $\square$ \\
Topic Relevance      & $\square$ & $\square$ & $\square$ & $\square$ & $\square$ & $\square$ & $\square$ & $\square$ & $\square$ & $\square$ \\
Fact Authenticity    & $\square$ & $\square$ & $\square$ & $\square$ & $\square$ & $\square$ & $\square$ & $\square$ & $\square$ & $\square$ \\
Logical Validity     & $\square$ & $\square$ & $\square$ & $\square$ & $\square$ & $\square$ & $\square$ & $\square$ & $\square$ & $\square$ \\
\bottomrule
\end{tabular}
\\[0.05em]

{\large \textbf{Scoring Table: Candidate 2}}\\[0.2em]
\begin{tabular}{lcccccccccc}
\toprule
\textbf{Dimension} & 1 & 2 & 3 & 4 & 5 & 6 & 7 & 8 & 9 & 10 \\
\midrule
Emotional Appeal     & $\square$ & $\square$ & $\square$ & $\square$ & $\square$ & $\square$ & $\square$ & $\square$ & $\square$ & $\square$ \\
Argument Clarity     & $\square$ & $\square$ & $\square$ & $\square$ & $\square$ & $\square$ & $\square$ & $\square$ & $\square$ & $\square$ \\
Argument Arrangement & $\square$ & $\square$ & $\square$ & $\square$ & $\square$ & $\square$ & $\square$ & $\square$ & $\square$ & $\square$ \\
Topic Relevance      & $\square$ & $\square$ & $\square$ & $\square$ & $\square$ & $\square$ & $\square$ & $\square$ & $\square$ & $\square$ \\
Fact Authenticity    & $\square$ & $\square$ & $\square$ & $\square$ & $\square$ & $\square$ & $\square$ & $\square$ & $\square$ & $\square$ \\
Logical Validity     & $\square$ & $\square$ & $\square$ & $\square$ & $\square$ & $\square$ & $\square$ & $\square$ & $\square$ & $\square$ \\
\bottomrule
\end{tabular}
\\[0.05em]

{\large \textbf{Scoring Table: Candidate 3}}\\[0.2em]
\begin{tabular}{lcccccccccc}
\toprule
\textbf{Dimension} & 1 & 2 & 3 & 4 & 5 & 6 & 7 & 8 & 9 & 10 \\
\midrule
Emotional Appeal     & $\square$ & $\square$ & $\square$ & $\square$ & $\square$ & $\square$ & $\square$ & $\square$ & $\square$ & $\square$ \\
Argument Clarity     & $\square$ & $\square$ & $\square$ & $\square$ & $\square$ & $\square$ & $\square$ & $\square$ & $\square$ & $\square$ \\
Argument Arrangement & $\square$ & $\square$ & $\square$ & $\square$ & $\square$ & $\square$ & $\square$ & $\square$ & $\square$ & $\square$ \\
Topic Relevance      & $\square$ & $\square$ & $\square$ & $\square$ & $\square$ & $\square$ & $\square$ & $\square$ & $\square$ & $\square$ \\
Fact Authenticity    & $\square$ & $\square$ & $\square$ & $\square$ & $\square$ & $\square$ & $\square$ & $\square$ & $\square$ & $\square$ \\
Logical Validity     & $\square$ & $\square$ & $\square$ & $\square$ & $\square$ & $\square$ & $\square$ & $\square$ & $\square$ & $\square$ \\
\bottomrule
\end{tabular}
\\[0.05em]

{\large \textbf{Scoring Table: Candidate 4}}\\[0.2em]
\begin{tabular}{lcccccccccc}
\toprule
\textbf{Dimension} & 1 & 2 & 3 & 4 & 5 & 6 & 7 & 8 & 9 & 10 \\
\midrule
Emotional Appeal     & $\square$ & $\square$ & $\square$ & $\square$ & $\square$ & $\square$ & $\square$ & $\square$ & $\square$ & $\square$ \\
Argument Clarity     & $\square$ & $\square$ & $\square$ & $\square$ & $\square$ & $\square$ & $\square$ & $\square$ & $\square$ & $\square$ \\
Argument Arrangement & $\square$ & $\square$ & $\square$ & $\square$ & $\square$ & $\square$ & $\square$ & $\square$ & $\square$ & $\square$ \\
Topic Relevance      & $\square$ & $\square$ & $\square$ & $\square$ & $\square$ & $\square$ & $\square$ & $\square$ & $\square$ & $\square$ \\
Fact Authenticity    & $\square$ & $\square$ & $\square$ & $\square$ & $\square$ & $\square$ & $\square$ & $\square$ & $\square$ & $\square$ \\
Logical Validity     & $\square$ & $\square$ & $\square$ & $\square$ & $\square$ & $\square$ & $\square$ & $\square$ & $\square$ & $\square$ \\
\bottomrule
\end{tabular}
\\[0.1em]

{\large \textbf{Final Choice (pick one):}}\\[0.4em]
Candidate 1 $\square$ \quad Candidate 2 $\square$ \quad Candidate 3 $\square$ \quad Candidate 4 $\square$

\par\bigskip
\endgroup

\twocolumn
\clearpage

\end{document}